\documentclass[10pt,twocolumn,letterpaper]{article}

\usepackage[pagenumbers]{cvpr} 

\usepackage{graphicx}
\usepackage{amsmath}
\usepackage{amssymb}
\usepackage{booktabs}
\usepackage{qiangstyle}

\usepackage[pagebackref,breaklinks,colorlinks]{hyperref}

\usepackage[capitalize]{cleveref}
\crefname{section}{Sec.}{Secs.}
\Crefname{section}{Section}{Sections}
\Crefname{table}{Table}{Tables}
\crefname{table}{Tab.}{Tabs.}

\usepackage[dvipsnames]{xcolor}
\usepackage{xcolor}

\newcommand{\our}{\emph{FuseDream}}
\newcommand{\ganclip}{CLIP+GAN}
\newcommand{\ourloss}{\texttt{AugCLIP}}

\newcommand{\ourcomp}{\our\emph{-Composition}}

\def\cvprPaperID{0000} 
\def\confName{CVPR}
\def\confYear{2022}

\begin{document}

\title{\our: Training-Free Text-to-Image Generation \\with Improved CLIP+GAN Space Optimization} 

\author{
Xingchao Liu$^1$, Chengyue Gong$^1$, Lemeng Wu$^1$, Shujian Zhang$^1$, Hao Su$^2$, Qiang Liu$^1$\\
$^1$University of Texas at Austin, $^2$University of California, San Diego \\
\small{\texttt{\{xcliu, szhang19\}@utexas.edu, haosu@eng.ucsd.edu, \{cygong, lmwu, lqiang\}@cs.utexas.edu} }
}


\twocolumn[{%
\renewcommand\twocolumn[1][]{#1}%
\vspace{-1em}
\maketitle
\vspace{-1em}
\begin{center}
    \centering
    \vspace{-0.3in}
    \includegraphics[width=\linewidth]{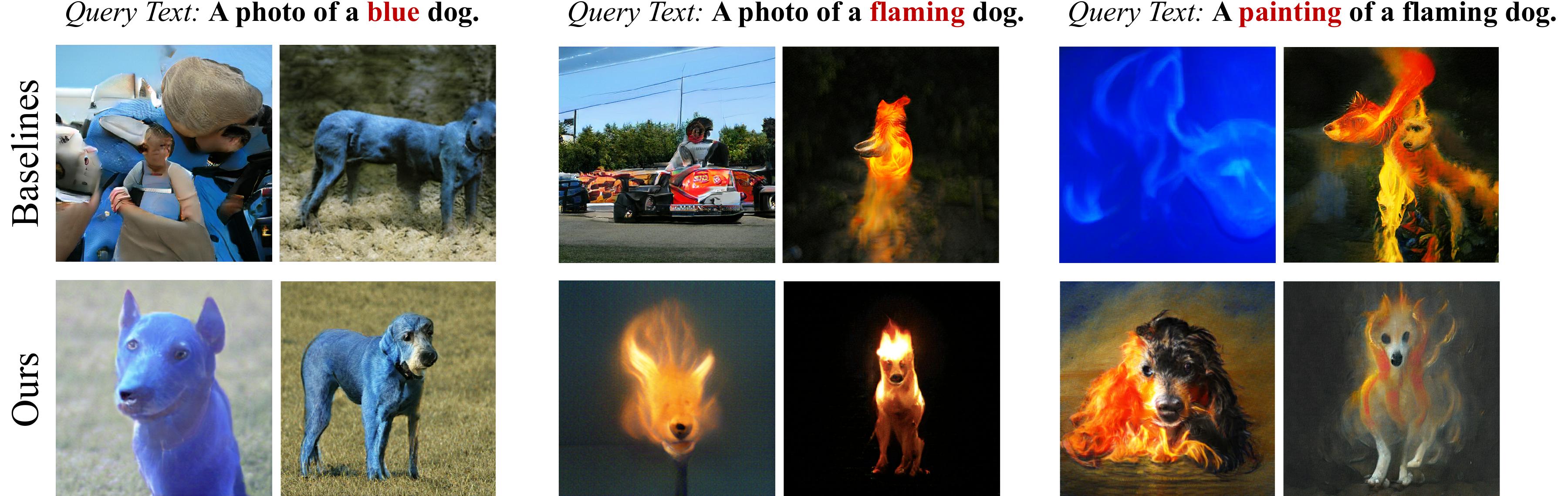}
    \captionof{figure}{Generated images for three query texts about dogs with different methods, including naive optimization with \ganclip~(top left of each panel), BigSleep~\cite{bigsleep} (top right of each panel), and \our~(the bottom row) with BigGAN-256 (left) and 512 (right). 
    }
    \label{fig:intro_dogs}
\end{center}
}]

\begin{abstract}
Generating images from natural language instructions is an intriguing yet highly challenging task. 
We approach text-to-image generation 
with a CLIP+GAN approach, which 
 optimizes in the latent space of an off-the-shelf GAN
 to find images that achieve maximum semantic relevance score with the given input text as measured by the CLIP model. 
Compared to traditional methods that train generative models mapping 
from text to image starting from scratch,  
the CLIP+GAN approach is training-free, zero-shot 
and can be easily customized with different generators. 

 However, 
optimizing CLIP score in the GAN space casts 
a highly challenging optimization problem and off-the-shelf optimizers such as Adam 
fail to yield satisfying results. 
In this work, we propose a {\our} pipeline, which improves the CLIP+GAN approach  
with three key techniques: 
1) a {\ourloss} score which robustifies the standard CLIP score by introducing random augmentation on image. 
2) a novel initialization and over-parameterization  strategy for optimization which allows us to efficiently navigate the non-convex landscape in GAN space. 
3) 
a \emph{composed generation} technique which, by leveraging a novel bi-level optimization formulation, can compose multiple images to extend the GAN space and overcome the data-bias.
 
When promoted by different input text, 
{\our} can generate high-quality images
with varying objects, backgrounds, artistic styles, 
 and novel counterfactual concepts that do not appear in the training data of the GAN that we use.  
Quantitatively, the images generated by {\our} yield top-level Inception score and FID score on MS COCO dataset, without additional architecture design or training.
Our code is publicly available at \url{https://github.com/gnobitab/FuseDream}.
\end{abstract}

\section{Introduction}





A landmark task in multi-modal machine learning 
is text-to-image generation, 
generating realistic images that are semantically related to a given text input 
\cite{reed2016generative, xu2018attngan, li2019object, tao2020df, ramesh2021zero, ding2021cogview}. 
This is a highly challenging task because 
the generative model needs to understand the text, image, and how they should be associated semantically. 
Recently, significant progresses have been achieved by 
\cite{ramesh2021zero, ding2021cogview} which generate high quality and semantically relevant images using models trained with self-supervised loss on large-scale datasets. 

The traditional approach to text-to-image generation is to train 
a conditional generative model from scratch with 
a dataset of $(\text{text, image})$ pairs~\cite{mansimov2015generating, reed2016generative, xu2018attngan, li2019object, tao2020df, ramesh2021zero}. 
This procedure, however, requires to collect a large training dataset, casts a high training cost, and can not be easily customized. 
Recently,
a more flexible text-to-image generation approach 
is enabled 
with the availability of powerful joint text-image encoders (notably the CLIP model \cite{radford2021learning}) that provide 
faithful semantic relevance score of text-image pairs.
Together with powerful pre-trained GANs (such as  \cite{abdal2019image2stylegan, brock2018large, li2019object, zhu2019dm}),
it is made possible to do text-to-image generation 
by optimizing in the latent space of a GAN to create images 
that have high semantic relevance with the input text. 
Notable examples include BigSleep \cite{bigsleep}
and VQGAN+CLIP \cite{VQGANCLIP}, which can generate intriguing and artistic images  
from text by maximizing the CLIP score in the latent space of BigGAN and VQGAN, respectively. 
Compared with the traditional benchmarks, 
methods that combine 
GAN and CLIP are training-free and zero-shot, 
requiring no dedicated training dataset and training cost.  
It is also much more flexible and modular: 
a user can easily replace the generator (GAN) or the encoder model (CLIP)
with more powerful or customized ones 
that fit best for their own problems and computational budget. 

On the other hand, the results from 
the existing {\ganclip} methods ~\cite{bigsleep, galatolo2021generating, VQGANCLIP}
are still not satisfying in many cases.   
For example, although BigSleep can generate 
images in different styles and create interesting visual arts, 
it finds difficulty in generating clear and realistic images and the resulting images can be only weakly related to the query text.
As shown in Figure~\ref{fig:intro_dogs} (top right of each panel),   
BigSleep can not generate a clearly recognizable image
for the simple concept of \emph{`blue dog'}.
For counterfactual concepts like \emph{`flaming dog'}, 
the image given by BigSleep tends to entangle the flame and dog concepts in an unnatural way. 
In Figure~\ref{fig:intro_dogs} (top left of each panel),
we  implement another baseline that  maximizes the CLIP score 
in the input space of BigGAN \cite{brock2018large} with the off-the-shelf Adam \cite{kingma2015adam} optimizer, which yields even worse results than BigSleep. 

In this work, 
we analyze the problems in the existing {\ganclip} procedures. 
We identity three key bottlenecks of the exiting approach and address them with a number of techniques to significantly improve the pipeline.  

    \paragraph{1) Robust Score} 
    We observe that the original CLIP score 
    does not serve as a good objective function for optimizing in the GAN space,
    as it tends to yield semantically unrelated images that ``adversarially'' maximize the CLIP score.  
    We propose an {\ourloss} score, 
    which robustifies the CLIP score  
    by averaging it on multiple perturbation or augmentation of the input images. 
    
    \paragraph{2) Improved Optimization Strategy} 
    Maximizing the CLIP score in the GAN space 
    yields a highly non-convex, multi-modal optimization problem and off-the-shelf optimization methods tend to be stuck at sub-optimal local maxima.
    We address this problem with a novel initialization and over-parameterization strategy which allow us to traverse in the non-convex loss landscape more efficiently.
    
\paragraph{3) Composed Generation} 
The image space of the {\ganclip} approach 
is limited by the pre-trained GAN that we use.   
This makes it difficult to generate images 
with novel combinations of objects that did not appear in the training data of the GAN. 
We address this problem by proposing a \emph{composed generation} technique, 
which co-optimizes two images so that they can be seamless composed together to yield a natural and semantically relevant image.  
We formulate composed generation into a novel bi-level optimization problem, 
which maximizes {\ourloss} score while 
incorporating a perceptual consistency score
as the secondary objective, and solve it efficiently with a recent dynamic barrier gradient descent algorithm \cite{gong2021automatic}.  

Our pipeline, which we call  \our\footnote{The "fuse" in the name refers to both the idea of 1) fusing GAN and CLIP and 2) our composed generation technique.},  
can generate not only clear objects {from} complex {text} description, but also complicate scenes {as these in MS COCO} \cite{lin2014microsoft}.  
Thanks to the representation power of CLIP, 
{\our} can create images with 
different backgrounds, textures, locations, artistic styles, and even  counterfactual objects. 
With the composed generation techniques, 
{\our} can create images with novel combinations of objects that do not appear in original training data of the GAN that we use. 
Comparing to directly training large-scale text-to-image generative models, 
our method is much more computation-friendly while achieving comparable or even better results.

\section{Text-to-Image Generation with \ganclip}
We first introduce the general idea of text-to-image generation 
by combining pre-trained image generators (in particular GANs)  
and joint image+text encoders (in particular CLIP).  
We then analyze a key limitation of 
 the naive realization of this approach. 

\paragraph{GAN} 
An image generator $g \colon \RR^D \to \RR^{H\times W \times 3} $ is a neural network that takes an $D$-dimensional latent code $\vv \xi$ and output a  colored image $\mathcal I$ of size $H\times W$. 
Formally, 
$$
\mathcal I = g(\vv \xi).  
$$
One can generate and manipulate different images by controlling the input $\vv \xi$. 
In this work, 
we use BigGAN \cite{brock2018large} 
unless otherwise specified, 
which is a class-conditional GAN whose latent vector $\vv \xi = \{\vv z, ~ \vv y\}$ includes both a Gaussian noise vector $\vec{z} \in \RR^Z$ 
and a class embedding vector $\vec{y} \in \RR^Y$.  
It was trained on the large-scale ImageNet dataset~\cite{russakovsky2015imagenet} with objects from $1,000$ different categories.  


\paragraph{CLIP}
A joint image-text encoder, 
notably \emph{Contrastive Language-Image Pretraining} (CLIP) ~\cite{radford2021learning}, 
consists of a pair of 
language encoder $\ftext$ and image encoder $\fimage$, 
which map a text $\mathcal{T}$ and an image $\mathcal I$ 
into a common latent space on which their relevance can be evaluated by cosine similarity 
\begin{equation}
    s_{\text{CLIP}}(\mathcal{T}, \mathcal{I}) = \frac{\left < \ftext(\mathcal{T}),~\fimage(\mathcal{I}) \right >}{||\ftext(\mathcal{T})||\cdot||\fimage(\mathcal{I})||}. 
\end{equation}
The CLIP model was trained such that 
semantically related  pairs of $\mathcal{T}$ and  $\mathcal{I}$
have high similarity scores. 

\paragraph{\ganclip}
One can synthesis a text-to-image generator by 
combining a pre-trained GAN $g$ and CLIP $
\{f_{text}, f_{image}\}$. 
Given an input text $\mathcal{T}$, 
we can generate a realistic image $\mathcal I$ that is semantically related to $\mathcal T$ by optimizing the latent code $\vv \xi$ such that the generated image $\mathcal I = g(\vv \xi)$ has maximum CLIP score $\sclip(\mathcal T, ~ \mathcal I)$. Formally, 
\begin{equation}
\label{eq:naive}
    \max_{\vec \xi}  
    \sclip\left (\mathcal{T}, g(\vec \xi) \right ).
\end{equation}
This confines the output image within the space of natural images 
while maximizing the semantic relevance to the input text. 
The optimization is 
solved with Adam~\cite{kingma2015adam} in \cite{zhu2016generative, bigsleep}. We truncate $\vec z$ to $[-2, 2]$ as a typical practice when using BigGAN~\cite{brock2018large, huh2020transforming}. 

\subsection{CLIP Can be Easily Attacked and Stuck}  
Naively solving \eqref{eq:naive} does not yield 
satisfying images
as shown in the top-left images of Figure~\ref{fig:intro_dogs}. 
We observe that the unsatisfying results  
can be attributed to two interconnected reasons:  

\noindent 1) CLIP scores can be easily ``attacked'', 
in that it is easy to maximize 
 CLIP  
within a small neighborhood 
of any image, indicating the existence of ``adversarial" 
images with high CLIP score but low semantic relevance with the input text. 

\noindent 2) The optimization in \eqref{eq:naive} can indeed effectively act as 
an adversarial optimization 
on $\sclip$, yielding images that are similar to the initialization but spuriously high CLIP score. 

\begin{figure}
    \centering
    \includegraphics[width=0.48\textwidth]{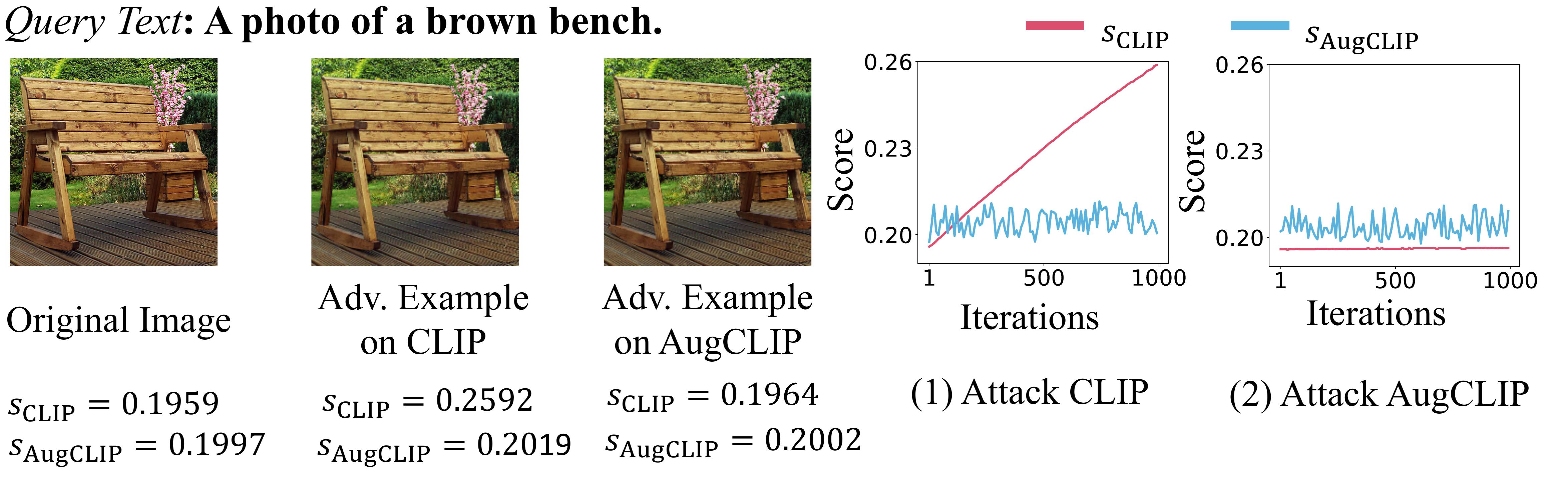}
    \caption{Results of adversarial attack on CLIP and AugCLIP score. (1) By applying FGSM on CLIP score, we get an image that has high $\sclip$ but visually identical to the original image. (2) Our \ourloss~score is robust against the FGSM attacking. 
    }
    \label{fig:clip_score}
\end{figure}

\paragraph{Case Study 1: Attacking CLIP} 
As shown in Figure~\ref{fig:clip_score}, 
we apply an adversarial  attacker, 
Fast Gradient Sign Method (FGSM)~\cite{goodfellow2014explaining} on $\sclip$ 
on a natural image $\mathcal{I}$, that is, we solve $\max_{\vv \delta} \sclip(\mathcal I + \vv \delta)$ $s.t.$ $\norm{\vv\delta} \leq \epsilon$ with a small perturbation magnitude $\epsilon >0$. 
We find that  FGSM can easily find an image that is almost identical with the original image, 
yet with much higher CLIP score. 
This indicates a danger of ``overfitting'' when we directly maximize the CLIP score.

\begin{figure}
    \centering
    \includegraphics[width=0.48\textwidth]{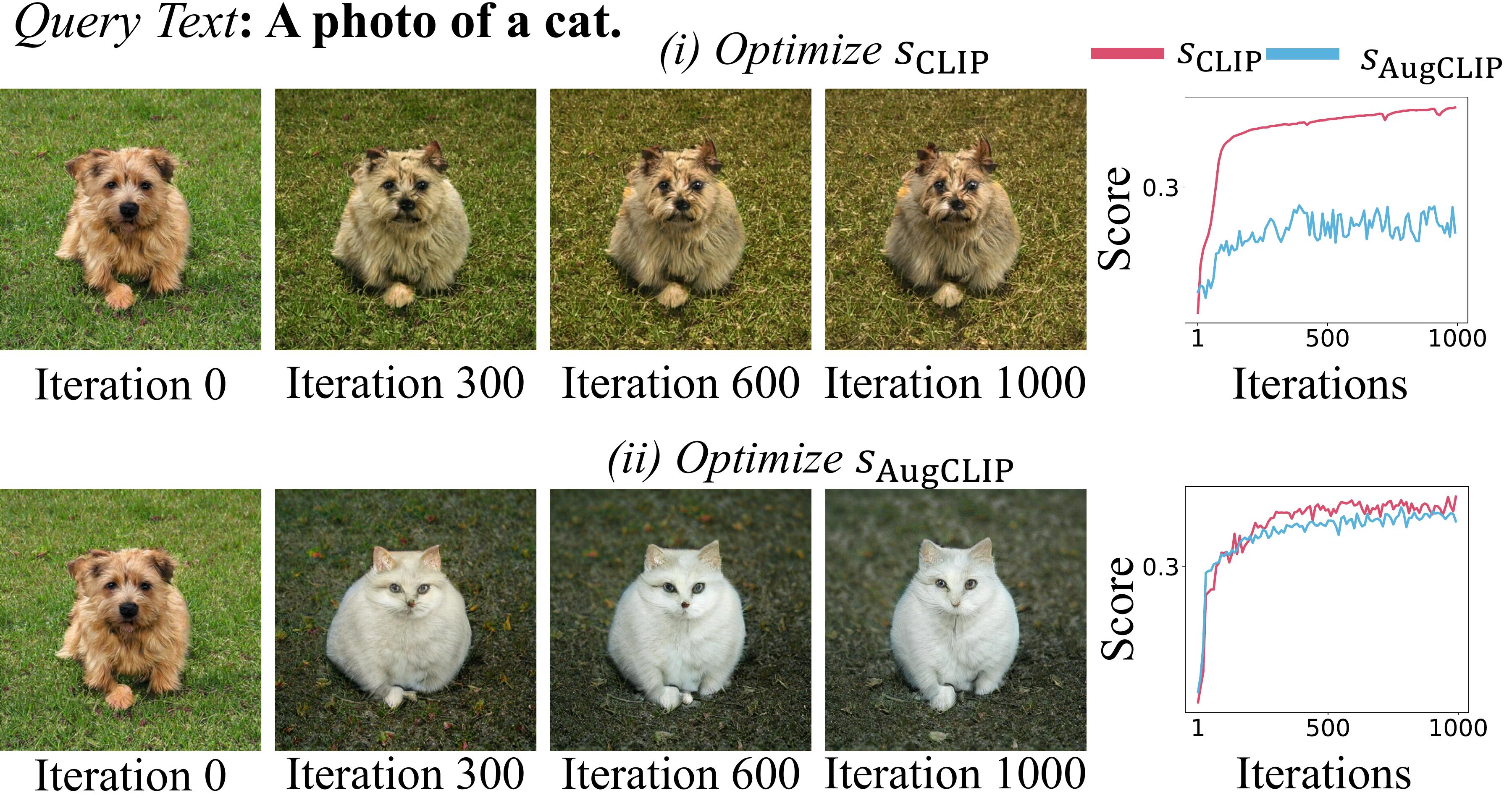}
    \caption{Optimizing $\sclip$ and $s_{\ourloss}$ in the GAN space as \eqref{eq:naive}.  
    Initialized from a dog image, 
   maximizing $\sclip$ only slightly changes in color of image while yielding spurious increase of $\sclip$ (top row). But maximizing $s_{\ourloss}$  quickly turns the dog into a white cat (bottom row). 
   Note that maximizing $s_{\ourloss}$ yields a similar amount of increase on $\sclip$, but not the other way around. 
    }
    \label{fig:augclip_score}
\end{figure}

\paragraph{Case Study 2: Dog$\to$Cat}
In Figure~\ref{fig:augclip_score}, 
we show an example of optimizing \eqref{eq:naive} 
with an input text $\mathcal{T}$=\emph{`A photo of a cat'}, 
initialized from an $\vv \xi^0$
whose image $\mathcal I = g(\vv\xi^0)$ is a dog. 
We can see that although   $\sclip$ is successfully maximized, 
the image remains similar to the initialization and 
does not transfer from a dog to a cat as expected.  
In this case, the
optimization in \eqref{eq:naive} exhibit  an adversarial attacking like behavior: it is stuck nearby the initialization while spuriously increasing the CLIP score.  

In both cases above, the problem can be resolved by using our {\ourloss} score, which we introduce in sequel. 






\section{Our Method - \our}
We now introduce our main techniques 
for improving the \ganclip~pipeline. 
Section~\ref{subsec:augclip} 
introduces the {\ourloss} score which robustifies 
the CLIP score to avoid the adversarial attacking phenomenon. 
Section~\ref{subsec:strategy} 
introduces an initialization and over-parameterization technique 
to better solve the non-convex optimization. 
Section~\ref{subsec:fusion} 
introduces a \emph{composed generation} to generate out-of-distribution images 
with novel combination of objects and backgrounds. 

\subsection{AugCLIP: Avoiding Adversarial Generation}
\label{subsec:augclip}
To address the adversarial attacking issue of the CLIP score, 
we propose the following {\ourloss} score, 
\begin{equation}
\label{eq:augclip}
    s_{\ourloss}(\mathcal{T}, \mathcal{I}) = \E_{\mathcal I' \sim \pi(\cdot|\mathcal I)} \left [\sclip(\mathcal{T},~ \mathcal{I}') \right],
\end{equation}
where $\mathcal I'$ is a random perturbation 
of the input image $\mathcal I$ drawn from a distribution $\pi(\cdot|\mathcal I)$ of candidate data augmentations. 
In our work, we adopt the various data augmentation techniques considered in DiffAugment~\cite{zhao2020differentiable},  
including random colorization, random translation, random resize, and random cutout.

{\ourloss} is more robust against adversarial attacks, because a successful adversarial perturbation against {\ourloss} must simultaneously attack $\sclip$ on most of the randomly augmented images, which is must harder than attacking a single image. The averaging over random augmentation also makes the  landscape smoother and hence harder to attack, as shown theoretically and empirically in~\cite{cohen2019certified, salman2019provably}.
Meanwhile, adding the augmentation  
does not hurt the semantic relation encoded by CLIP, because 
the CLIP model was originally trained on images with different colorizations, views and translations, and is hence compatible with our augmentation strategy. 

\begin{figure*}
    \centering
    \begin{tabular}{c} 
    \includegraphics[width=0.98\textwidth]{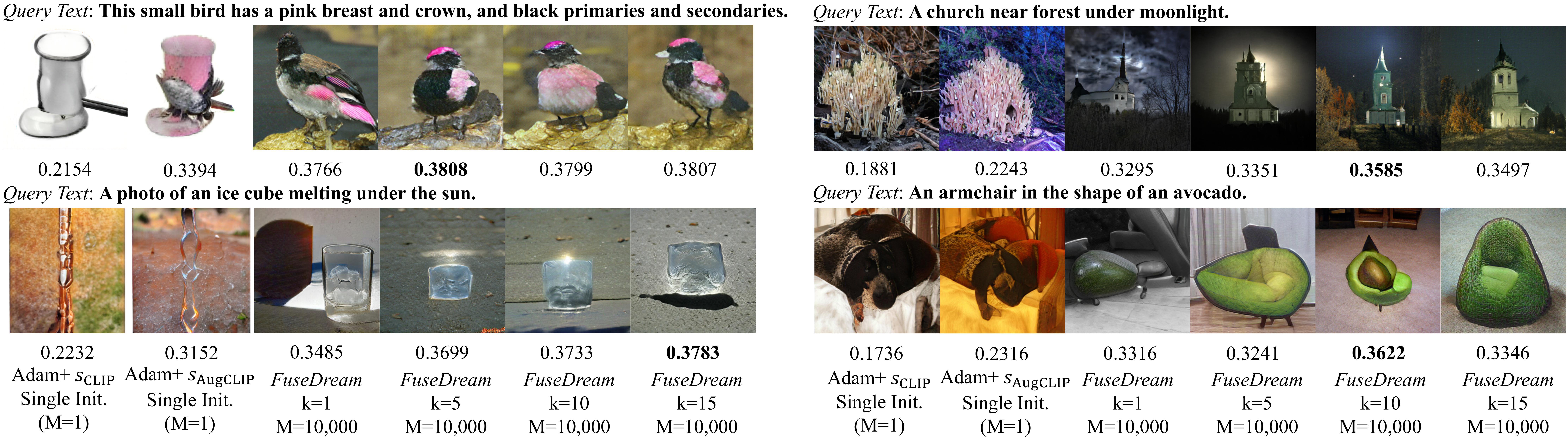} \\
    \vspace{-15pt}
    \end{tabular}
\caption{
Ablation of different initialization and optimization strategies (Init.=Initialization). 
The numbers below the images are the  $s_{\ourloss}$ score.
$\vec z$ is randomly initialized from the standard Gaussian distribution and $\vec y$ is initialized from the latent representations from the 1,000 classes in ImageNet.
The left top query text is from the CUB dataset~\cite{welinder2010caltech}. 
We can see that: 
(1) images with higher $s_{\ourloss}$  tend to  interpret the input text better,  indicating the effectiveness of $s_{\ourloss}$; 
(2) {\our} (right three columns of each panel) 
with $M=10,000$ and large $k$ generates high $s_{\ourloss}$ and high-quality images with multiple mixed concepts and nonexistent objects.
}
    \label{fig:opt}
\end{figure*}

\paragraph{Case Study 1 \& 2}
As shown in Figure~\ref{fig:augclip_score}, 
the {\ourloss} 
score is significantly more robust against adversarial attacking. 
 Figure~\ref{fig:opt} shows that simply replacing $\sclip$ with $s_{\ourloss}$ allows us to escape the adversarial generation regime and yield more semantically relevant images. 

\subsection{Improving Optimization}
\label{subsec:strategy}
The optimization 
of $s_{\ourloss}$ can still suffer from sub-optimal local maxima 
due to the high non-convexity of loss landscape. 
We introduce 
an initialization and over-parameterization strategy to 
  improve the optimization.  


Unlike traditional methods that start from a single initialization,
we start with sampling a large number $M$ of copies of initialization  $\{\vv \xi_i^0\}_{i=1}^M$. We then select the top $k$ initialization, say $\{\vv \xi_{(i)}^0\}_{i=1}^k$, that has the highest $\ourloss$ score,  
and use them as the \emph{initial basis vector} for the subsequent optimization, that is, we reparameterize the solution into $\vv \xi = \sum_{i=1}^k w_{(i)} \vec \xi_{(i)} $ and jointly optimize the basis vectors $\{\vv \xi_{(i)}\}_{i=1}^k$ and the coefficients $\{w_{(i)}\}_{i=1}^k$ with $w_{(i)} \in \RR$: 
\begin{align}
    \label{eq:final_loss}
    \max_{\{\vv \xi_{(i)},  w_{(i)} \}_{i=1}^k} s_{\ourloss} \left (\mathcal{T}, g\left (\sum_{i=1}^k w_{(i)} \vec \xi_{(i)} \right) \right ), 
\end{align}
with $\{\vv \xi_{(i)}\}$ initialized from 
the  $k$ selected $\{\vv \xi_{(i)}^0\}$, and $w_{(i)}$ initialized from $1/k$. 
We set $M=10,000$ (which can be parallel evaluated)  and a relatively small $k$ (e.g., $k \leq 15$).
 
 Although the optimization in \eqref{eq:final_loss} is equivalent to that of \eqref{eq:naive}, 
 it is equipped with an 
 over-parameterized and 
 more natural coordinate and better initialization, 
 and hence tends to yield better results 
 when solved with gradient-based optimization methods. 
  In particular, 
 the update of the combination weights $\{w_{(i)}\}$ 
 corresponds to fast and global move in 
 the linear span  
 of the basis vectors $\{\vv \xi_{(i)}^0\}_{i=1}^k$, 
 making it easier to escape local optima. 
 
In practice, as we use BigGAN, the latent code $\vv \xi = (\vec z, \vec y)$ is initialized with  $\vv z\sim \normal(0,I)$ and  $\vec y$ randomly selected from the latent representations for the 1,000 classes of ImageNet 
(which is better than initializing $\vv y \sim \normal(0,I)$ as we show in Appendix). 

\paragraph{Gradient-based or Gradient-free Optimizer?} 
In this work, we adopt the widely-used Adam~\cite{kingma2015adam} optimizer.
Some recent works recommended to use 
gradient-free optimizers, 
like BasinCMA~\cite{wampler2009optimal, bau2019seeing, huh2020transforming}, 
for optimizing in GAN spaces due to the high non-convexity. 
However, our study shows  that BasinCMA 
tends to cast a higher computation cost than Adam,  because  BasinCMA requires  a large number of forward passes 
on the objective at each iteration, while Adam only requires a single forward and backward pass. 
Empirically, 
we found that Adam is $\times$20 faster than BasinCMA.
Although gradient-based methods 
are more prune to local optima than gradient-free methods, 
it becomes less an issue with our {\ourloss} loss and the proposed initialization and over-parameterization technique. 
We include more discussion and comparison with BasinCMA in  Appendix.

\begin{algorithm}[t]
\caption{{\our} (with single image generation)} 
\label{alg:single}
\begin{algorithmic}[1]
\STATE \textbf{Input:} Query text $\mathcal{T}$; 
a GAN $g$; CLIP $\{\ftext, \fimage\}$.  
\STATE Generate $M$ initialization $\{\vv\xi_i^0\}_{i=1}^M$ randomly, 
and  select the top-$k$  $\{\vv\xi_{(i)}^0\}_{i=1}^k$ with the highest 
 $s_{\ourloss}$. 
\STATE Solve 
the optimization problem~\eqref{eq:final_loss}
using Adam,  with $\{\vv\xi_{(i)}\}_{i=1}^k$ initialized from $\{\vv\xi_{(i)}^0\}_{i=1}^k$. 
\RETURN  image $\mathcal{I} = g\left (\sum_{i=1}^k w_{(i)} \vv\xi_{(i)}\right)$.
\end{algorithmic}
\end{algorithm}

\subsection{ Composed generation} 
\label{subsec:fusion}

\begin{figure*}
    \centering
    \includegraphics[width=0.98\textwidth]{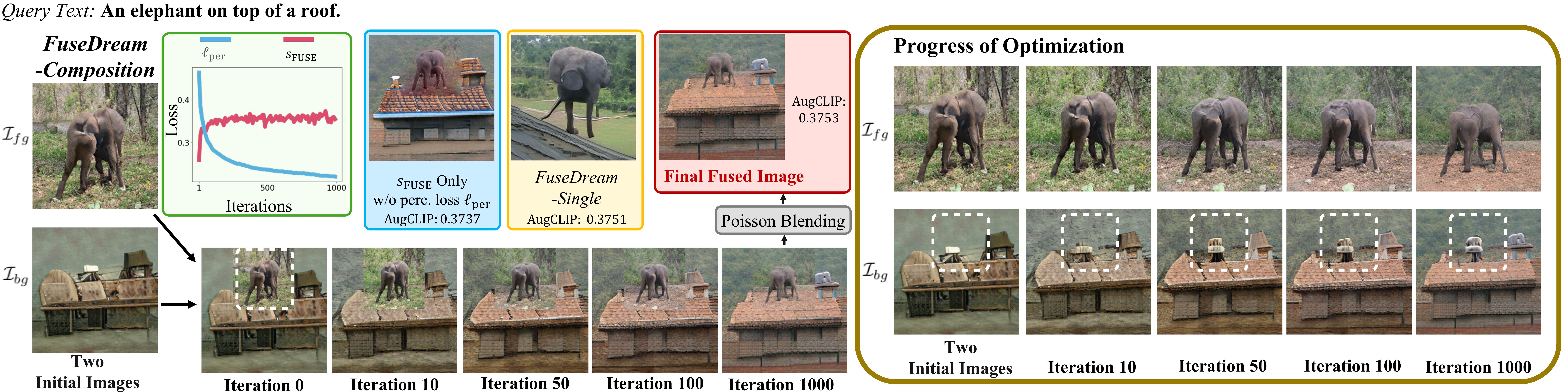}
    \caption{\textcolor{black}{Red box}: Image generated by {\ourcomp} on  \emph{`An elephant on top of a roof'}. 
    \textcolor{black}{Blue box}: Image given by composed generation that only optimizes the {\ourloss} score $s_{\fuse}$ without considering the perceptual loss $\ell_{\fuse}$,  
    which shows a clear discontinuity between $\mathcal{I}_{bg}$ and $\mathcal{I}_{fg}$. 
    \textcolor{black}{Yellow box}: Image given by {\our} without composition, 
    which has a less clear elephant without nose and unrecognizable `roof', 
    because BigGAN cannot handle composed concepts well.
    \textcolor{black}{Green Box}: The {\ourloss} score $s_{\fuse}$ and perceptual loss $\ell_{\fuse}$ vs. iteration using the algorithm in \eqref{eq:lexico_update}.  
    Note that the generated images in blue and red boxes have similar {\ourloss},  indicating that the bi-level optimization 
    can improve $\ell_{\fuse}$ for free without scarifying $s_{\fuse}$. 
     The white dashed box on $\mathcal{I}_{bg}$ indicates the position and size of $\mathcal{I}_{fg}$.  
    Notice how the style and the color of $\mathcal{I}_{fg}$ and $\mathcal{P}_{bg}$ slowly unify across the iteration. 
    }
    \label{fig:lexico}
    \vspace{-10pt}
\end{figure*}

\begin{algorithm}[t]
\caption{{\ourcomp}}
\label{alg:fuse}
\begin{algorithmic}[1]
\STATE \text{Input:} Query text $\mathcal{T}$; 
a GAN $g$; CLIP $\{\ftext, \fimage\}$. 
A finite candidate set $\Gamma$ 
of the composition parameter $\bar{\vv\alpha}$. 
\FOR{$\bar{\vv\alpha}$ in $\Gamma$}
\STATE Solve the optimization problem~\eqref{eq:bilevel} with \eqref{eq:lexico_update}, yielding a composed image   $\mathcal{I}_{\bar{\vv\alpha}}$.
\ENDFOR
\RETURN image $\mathcal{I}_{\bar{\vv\alpha}}, \bar{\vv\alpha}\in \Gamma$ with the highest $s_{\text{AugCLIP}}$.
\end{algorithmic}
\end{algorithm}

The space of images of the {\ganclip} approach is limited by the representation power of the GAN we use. 
This makes the method difficult to generate out-of-distribution images and 
prune to inherit the data biases 
 e.g., center, spatial and color bias \cite{anirudh2019mimicgan, huh2020transforming}, 
 from the original training set of the GAN. 
We propose \emph{composed generation}, 
which expands the image space and reduce the data bias by composing two images generated by GAN to gain highly flexibility.  

Our method co-optimizes a 
foreground image $\mathcal{I}_{fg}=g(\bar {\vv{\xi}}_{fg})$ 
and a
background image $\mathcal{I}_{bg}=g(\bar{\vv\xi}_{bg})$,
where $\bar{\vv{\xi}}_{fg}$ and $\bar{\vv{\xi}}_{bg}$ are two over-parameterized latent codes as Eq~\ref{eq:final_loss}. 
The two images are used to generate a fused image 
$$\mathcal I_{\fuse}
=\fuse(\mathcal I_{fg}, \mathcal I_{bg}, \{\alpha, \vv t\})$$ by first scaling $\mathcal I_{fg}$ in size with a factor $\alpha\in(0,1)$, and then pasting it on top of $\mathcal I_{bg}$ at one of 9 locations $\vv t \in\{\textrm{left, center, right}\}^2$. We hope to choose $\bar{\vv\xi}:=\{\bar{\vv \xi}_{fg},\bar{ \vv \xi}_{bg}\}$, and $\bar{\vv\alpha}:=\{\alpha, \vv t\}$
to maximize the {\ourloss} score of $\mathcal I$: 
$$
s_{\texttt{Fuse}}(\bar{\vv \xi},\bar {\vv\alpha })
:= s_{\ourloss}(\mathcal T,\mathcal I_\fuse). 
$$
 %
On the other hand, 
because the two images $\mathcal{I}_{fg}$ and $\mathcal{I}_{bg}$ 
are generated independently,
the composed image may have unnatural and  artificial discontinuity on the boundary. To address this, we introduce an additional loss that enforces the perceptual consistency between $\mathcal{I}_{fg}$ and $\mathcal{I}_{bg}$,
$$
\ell_{\fuse}(\bar{\vv \xi}, \bar{\vv\alpha}): = \ell_{\texttt{per}}(\mathcal I_{fg}, \texttt{Crop}(\mathcal I_{bg}, \{\alpha, \vv t\})),
$$
where $\ell_{\texttt{per}}$ denotes the LPIPS metric~\cite{zhang2018unreasonable}, 
a measure 
that approximates human perception of image similarity. 
Hence, we want to both maximize  the {\ourloss} score and minimize the perceptual loss $\ell_{\fuse}$. 
A naive approach is to optimize their linear combination. 
However, this would require a careful and case-by-case tuning of the combination coefficient when generating each image. 

\paragraph{Bi-level Optimization}
We propose a  
\emph{tuning-free} approach for combining two losses via 
a simple bi-level (or lexicographic) optimization problem (see e.g., \cite{dempe2020bilevel,gong2021biobjective}), 
\begin{equation}
\label{eq:bilevel}
\begin{aligned}
    & \min_{\bar{\vv \xi}, \bar{\vv\alpha}} \ell_{\fuse}(\bar{\vv \xi}, \bar {\vv \alpha})
    & s.t. ~~~ (\bar{\vv \xi}, \bar{\vv \alpha}) \in  
    \argmax  s_{\fuse},
\end{aligned}
\end{equation}
where   $\argmax  s_{\fuse}$ denotes the set of (local) maxima of $s_{\fuse}$. 
This formulation seeks in the optimum set of $s_{\fuse}$ the points that minimize $\ell_{\fuse}$. It prioritizes the optimization of $ s_{\fuse}$ while incorporating $\ell_{\fuse}$ as a secondary loss. 

We optimize $\bar{\vv \alpha} = \{\alpha, \vv t \}$ by brute-force search in the discrete set $\alpha \in\{0.65, 0.5\}$ and $\vv t \in \{\textrm{left, center, right}\}^2$. For each fixed $\bar{\vv\alpha}$, we optimize the continuous vector $\bar{\vv\xi} $ using the dynamic barrier gradient descent algorithm from \cite{gong2021biobjective},  
which yields the following simple update rule at iteration $t$, 
\begin{equation}
\begin{aligned}   
    \label{eq:lexico_update}
    &\bar{\vv\xi}^{t+1} \leftarrow \bar{\vv\xi}^{t} - \epsilon^t \vv v^t,   
    ~~~~~ \vv v^t = \nabla 
    \ell_{\fuse}^t  - \lambda_t \nabla 
    s_{\fuse}^t,
    \\ &\lambda_t = \max  \left (\frac{\beta ||\nabla
    s_{\fuse}^t||^2 + 
    \langle \nabla \ell_{\fuse}^t,~ \nabla
    s_{\fuse}^t\rangle }{||\nabla
    s_{\fuse}^t ||^2},~0 \right ),
\end{aligned}
\end{equation}
where $\epsilon^t >0$ is the step size; $\beta$ is a hyperparameter ($\beta=1$ by default); 
we wrote that
$\nabla s_{\fuse}^t = \nabla_{\bar{\vv\xi}}
    s_{\fuse}(\bar{\vv\xi}^t)$  and 
$\nabla \ell_{\fuse}^t = \nabla_{\bar{\vv\xi}}
    \ell_{\fuse}(\bar{\vv\xi}^t)$. 
    
    Intuitively, one may view this algorithm as iteratively minimizing the linearly combined loss $\ell_{\fuse} - \lambda_t s_\fuse$, with the coefficient  $\lambda_t$ dynamically decided by the angle between gradient $\dd \ell_{\fuse}$ and $\dd s_{\fuse}$, in a way that removes the component of $-\dd \ell_\fuse$ that is conflict with $\dd s_{\fuse}$, so that 
    $s_\fuse$ is ensured to decrease monotonically given that it is the primary loss.   
See Appendix and \cite{gong2021automatic}  for more details. 

In practice, we combine \eqref{eq:lexico_update} with Adam by treating $\vv v^t$ as the gradient direction. 
In addition, we obtain the final composed image by applying  Poisson blending on $\mathcal{I}_{fg}$ and $\mathcal{I}_{bg}$ to yield even smoother image $\mathcal{I}$ following ~\cite{huh2020transforming}. 
Our algorithm is summarized in Alg.~\ref{alg:fuse}.


\section{Related Works}
The general idea of optimizing in the latent space of GAN has been widely used as a powerful framework 
for generating, editing and recovering images; see, e.g.,  
~\cite{zhu2016generative, harkonen2020ganspace, huh2020transforming, abdal2019image2stylegan}, to name only a few.  
For example, \cite{zhu2016generative} proposes to project real images to the latent space of GAN to edit images. 
\cite{harkonen2020ganspace} applied principal component analysis to the GAN space 
to create interpretable controls for image synthesis. 
\cite{huh2020transforming} optimizes the latent code to embed a given image into the BigGAN~\cite{brock2018large} by a gradient-free optimizer, BasinCMA~\cite{bau2019seeing, wampler2009optimal}, to allow flexible image editing in the GAN space. 
A recent work~\cite{daras2021intermediate} uses a layer-wise optimization to improve the performance of solving inverse problems, such as super-resolution and inpainting, with GAN space.
Most of these methods solely focus on a single task on the image domain, while our method aims to connect images with text by leveraging the power of CLIP. 

On the other direction, the idea of using CLIP score~\cite{radford2021learning} has been explored in various directions, 
including video retrieval~\cite{luo2021clip4clip}, visual question answering~\cite{shen2021much}, and language-guided image manipulation/generation
~\cite{galatolo2021generating, bigsleep, VQGANCLIP, patashnik2021styleclip}. 
In particular, \cite{patashnik2021styleclip} adopts CLIP and StyleGAN to guide the style of a simple image, typically a photo of face, pet or car.  
\cite{bigsleep, VQGANCLIP, galatolo2021generating} are open-source repositories that implemented the vanilla GAN+CLIP procedure, which we improves substantially with the new techniques. 


\section{Experiments}

\begin{figure*}[t]
\centering
\scalebox{0.95}{
\begin{tabular}{c}
\includegraphics[width=1.\textwidth]{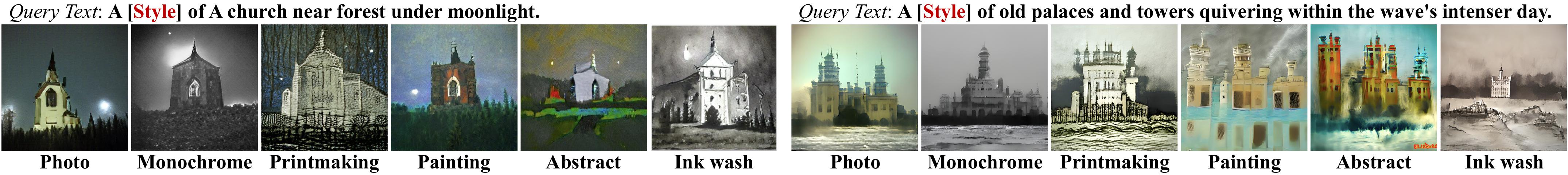}
\\
\includegraphics[width=1.\textwidth]{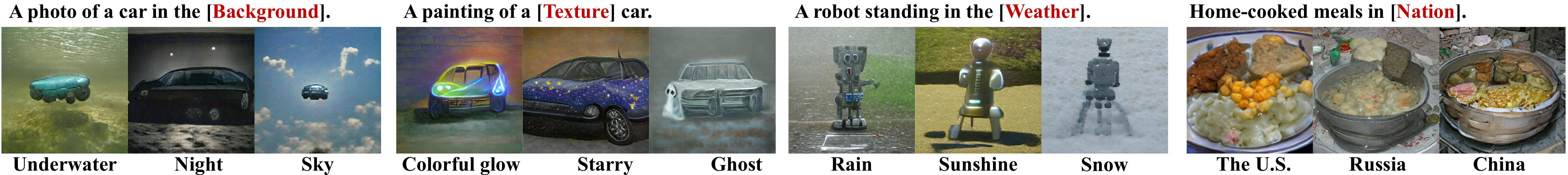}
\\
\includegraphics[width=1.\textwidth]{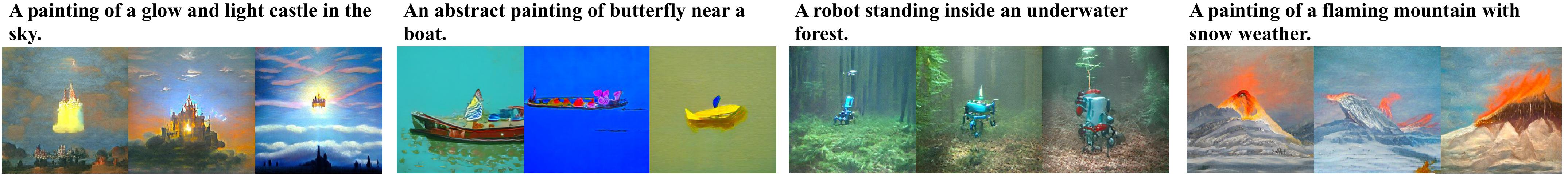}
\\
\end{tabular}}
\caption{First row: images with different styles generated by our method. Second row: images  with different backgrounds, textures, weather, and etc. Third row:   more complicated examples combining multiple objects, textures, styles, and  other information together. The three images in the third row are selected out of 5 random seeds according to the \ourloss~score.}
\label{fig:background_texture}
\vspace{-5pt}
\end{figure*}

\begin{figure}
    \centering
    \scalebox{0.45}{
    \begin{tabular}{c}
    \includegraphics[width=1.\textwidth]{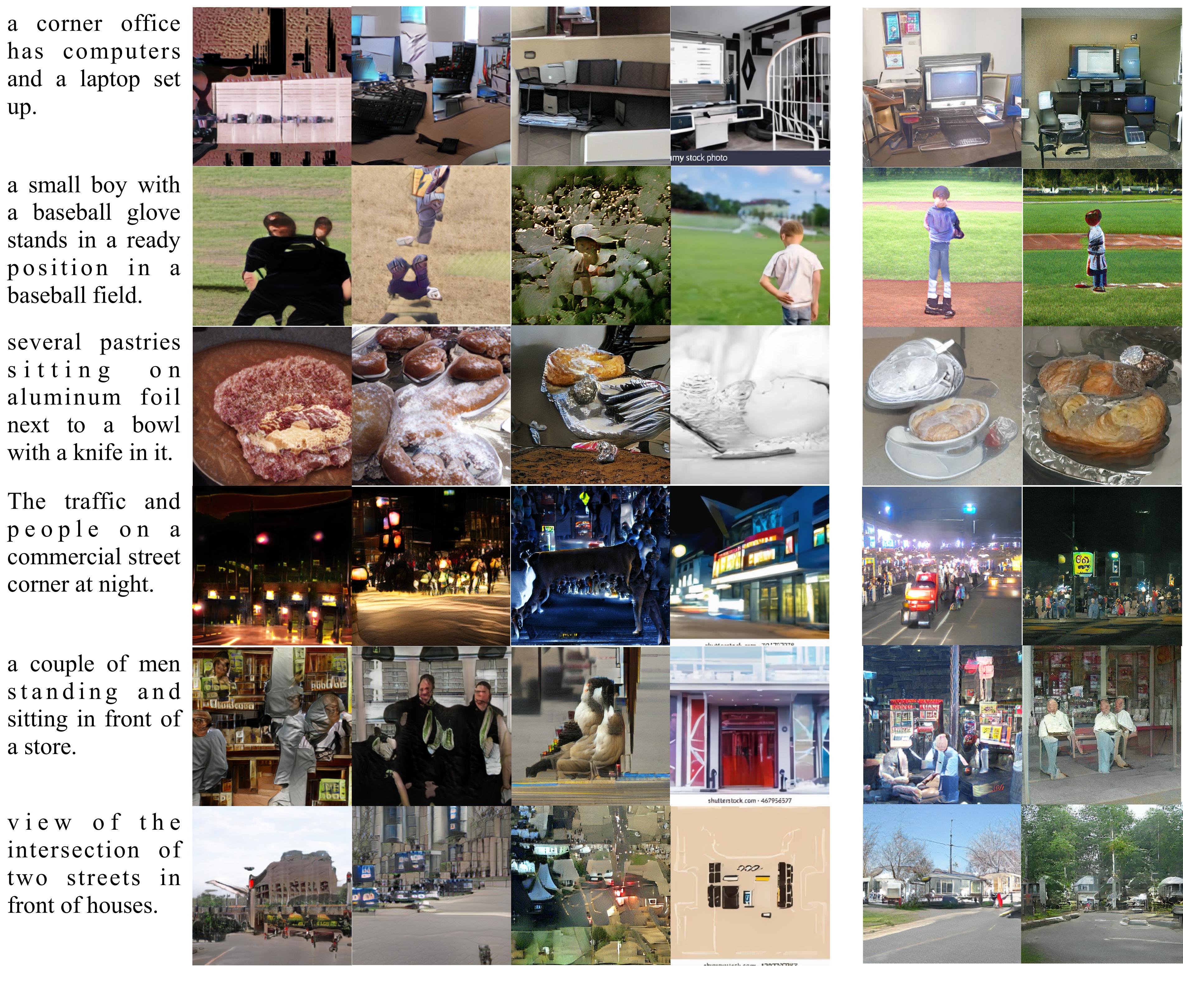}
    \vspace{-5pt}
    \\
~~~~~~~~~~~~~~~~~~~~~~~~~~~~~~~~~~~~~~~~~~ {DF-GAN} ~~~~~~~~~~~{DM-GAN} ~~~~~~~~
{BigSleep} ~~~~~~~~~~~~ {CogView} ~~~~~~~
{{\our}(256)}~~~~~{{\our}(512)}
    \end{tabular}}
    \caption{Comparison between \our~and DF-GAN~\cite{tao2020df}, DM-GAN~\cite{zhu2019dm}, BigSleep~\cite{bigsleep}, CogView~\cite{ding2021cogview} on MS COCO test set. \our~can handle the complicated captions from MS COCO and generate meaningful images.}
    \label{fig:coco}
\end{figure}

\begin{figure}[t]
\centering
\scalebox{0.95}{
\begin{tabular}{c}
\includegraphics[width=.48\textwidth]{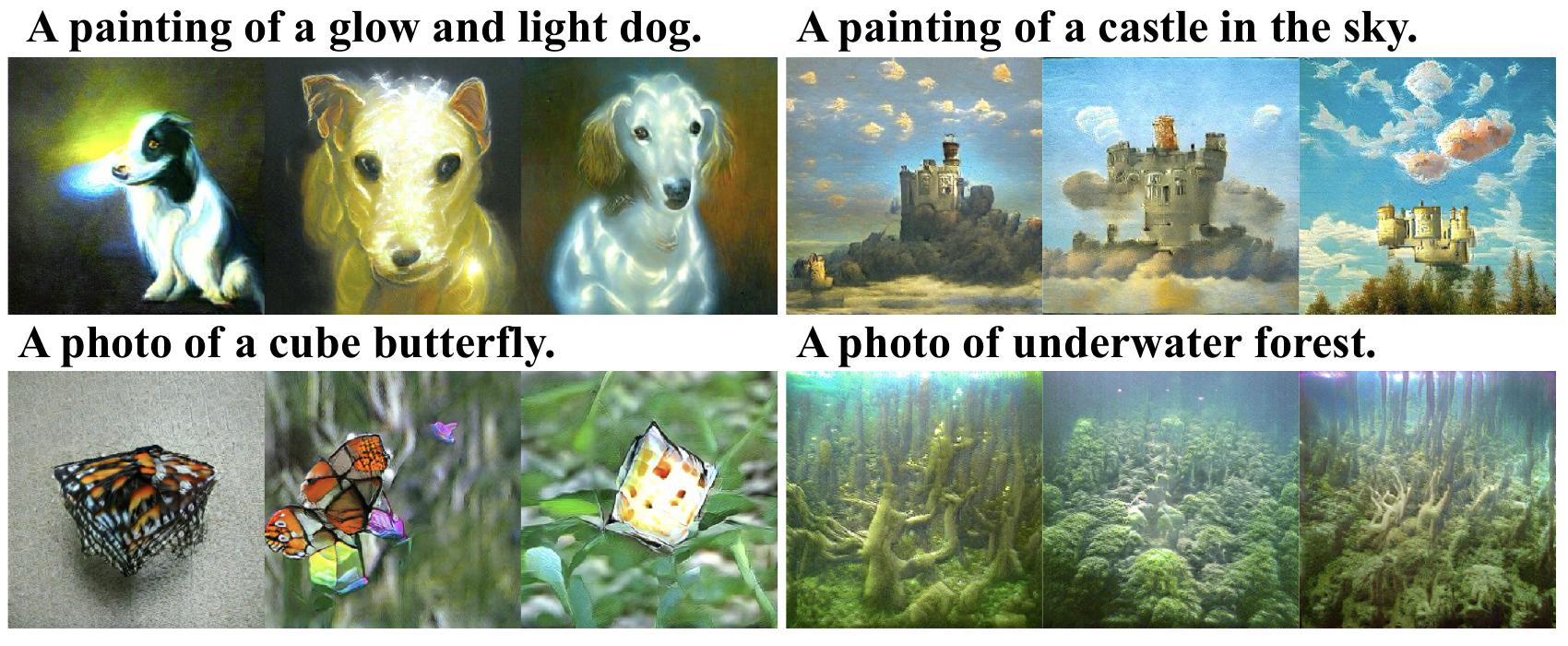}
\\
\end{tabular}}
\vspace{-5pt}
\caption{Counterfactual images generated by our method. The three images in each panel are selected out of 5 random seeds according to  \ourloss~value.}
\vspace{-10pt}
\label{fig:counterfactual}
\end{figure}

\begin{figure*}[t]
\centering
\scalebox{0.96}{
\begin{tabular}{c}
\includegraphics[width=1.\textwidth]{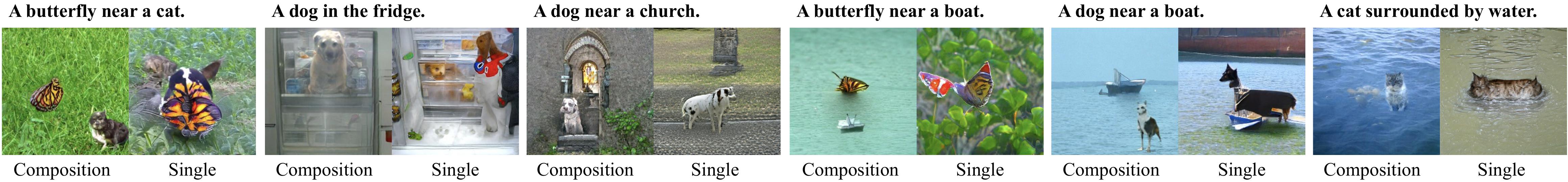}
\\
\includegraphics[width=1.\textwidth]{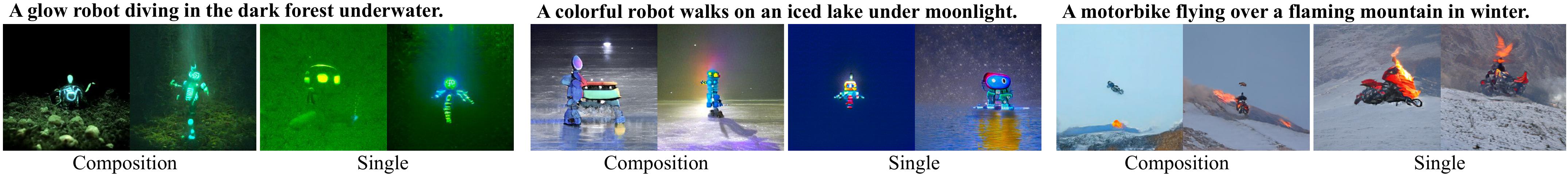}
\\
\end{tabular}}
\caption{
Images from {{\our} (with single image generation)} and {\ourcomp}. We find that: (1) {\ourcomp} can generate seamlessly fused images with no unnatural discontinuous on the fusing boundary; (2) Prompted with the simple query texts in the top row, {\our} (single) tends to mix two objects together or ignore some concepts in the query, while {\ourcomp} can generate images with clear and disentangled objects; (3) For the  more complex text in the second row, 
{\ourcomp} can generate images with fine-grained details (e.g., \emph{`dark forest'}, \emph{`walk'}, \emph{`moonlight'}, \emph{`flying'}, \emph{`flaming mountain'}).  
}
\label{fig:fusion}
\end{figure*}

\begin{figure*}[t]
\centering
\scalebox{0.96}{
\begin{tabular}{c}
\includegraphics[width=1.\textwidth]{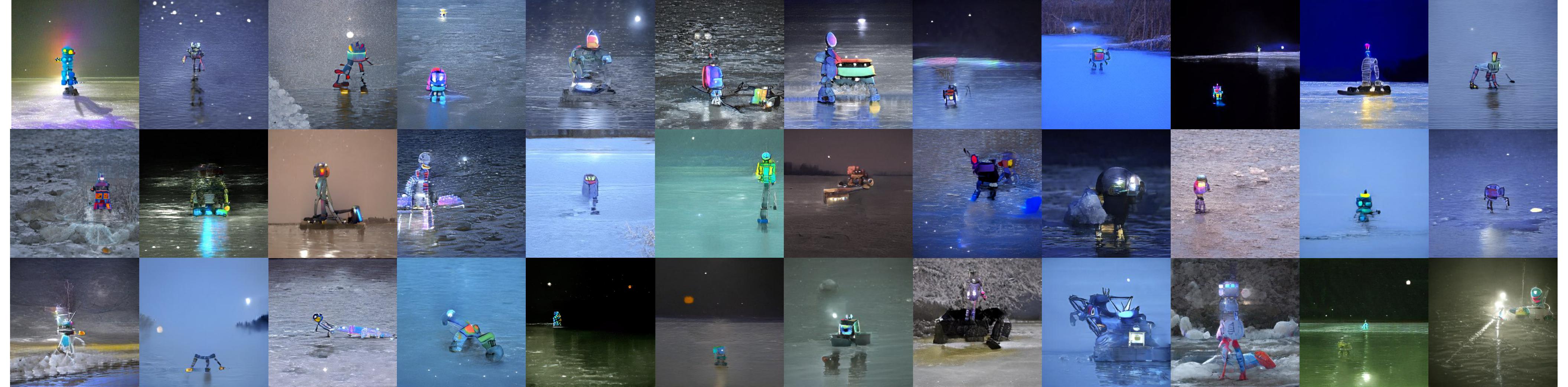}
\\
\end{tabular}}
\vspace{-5pt}
\caption{
More mages
generated by {\ourcomp} for \emph{`A colorful robot walks on an iced lake under moonlight'}, verifying that our method can generate high-quality images with different random seeds. The 36 images are
selected out of 50 random seeds according to the \ourloss~score, and are 
ordered with descending {\ourloss} score (from the left to right, the top to down).}
\label{fig:multiple_images}
\vspace{-5pt}
\end{figure*}

\begin{table}[t]
    \centering
    \scalebox{0.62}{
    \begin{tabular}{l|cccc}
    \hline
    \hline
    Methods    & IS ($\uparrow$) & FID ($\downarrow$) & CLIP R-prec ($\uparrow$) & R-prec ($\uparrow$) \\
    \hline 
    Real Images~\cite{hinzsemantic} & 34.88$\pm$0.01 & 6.09$\pm$0.05 & 82.84$\pm$0.04 & 68.58$\pm$0.08 \\
    \hline
    GAN-IC~\cite{reed2016generative} & 7.88$\pm$0.07 & -  & - & - \\
    StackGAN~\cite{zhang2017stackgan} & 8.45$\pm$0.03 & - & - &  - \\
    AttnGAN~\cite{xu2018attngan}   & 23.79$\pm$0.32 & 28.76   & 65.66$\pm$2.83 & 82.98$\pm$3.15     \\
    P-AttnGAN~\cite{li2019object} & 26.31$\pm$0.43 & 41.51 & - & 86.71$\pm$2.97 \\
    Obj-GAN~\cite{li2019object} & 27.37$\pm$0.22 & 25.85  & - & 86.20$\pm$2.98\\
    DM-GAN~\cite{zhu2019dm} & 32.32$\pm$0.23 & 27.34$\pm$0.11 & 65.45$\pm$2.18 & \textbf{91.87$\pm$0.28} \\
    OP-GAN~\cite{hinzsemantic} & 27.88$\pm$0.12 & 24.70$\pm$0.09 & - & 89.01$\pm$0.26 \\
    DF-GAN~\cite{tao2020df} & - & 21.42  &  66.42$\pm$1.49 & -  \\
    \hline
    DALL-E~\cite{ramesh2021zero} & 17.9 & 27.5 & - & -  \\
    CogView~\cite{ding2021cogview} & 18.2 & 27.1 & - & - \\
    BigSleep~\cite{bigsleep}  & 13.32 $\pm$ 0.45  & 66.46 & 91.89$\pm$0.31 & 28.75$\pm$3.32 \\ 
    \hline
    \textbf{\our~ (k=5) (256)}  & 34.26 $\pm$ 0.76 & \textbf{21.16}  &  96.43$\pm$0.24 & 65.56$\pm$1.37\\
    \textbf{\our~ (k=10) (256)}  & \textbf{34.67 $\pm$ 0.97} & 21.89 & \textbf{98.46$\pm$0.17}  & 66.06$\pm$0.59  \\
    \textbf{\our~ (k=5) (512)}  & 34.19$\pm$0.95  & 21.52 & 98.38$\pm$0.18 & 63.67$\pm$1.90 \\
    \textbf{\our~ (k=10) (512)}  & 32.88$\pm$0.93 & 25.24  &  \textbf{98.44$\pm$0.15} & 63.80$\pm$1.12  \\
    \hline
    \hline
    \end{tabular}}
    \caption{Comparison with  state-of-the-art text-to-image generators on the test set of MS COCO. 
    Our method gets the highest Inception Score and the lowest FID score.  
    Note that the text-to-image GANs in the second block are directly trained on COCO to maximize the retrieval performance of the model  in~\cite{xu2018attngan} during training, and hence have much higher R-precision even than the real images. 
    For fair comparison, we also report R-precision calculated by the CLIP model.  
    }
    \label{tab:coco}
\end{table}
We  compare {\our} equipped with BigGAN-256 with a number of baselines including DM-GAN \cite{zhu2019dm}, Obj-GAN \cite{li2019object}, CogView \cite{ding2021cogview}, and etc. 
We test the methods 
on the popular MS COCO dataset~\cite{lin2014microsoft}, and find that \our~clearly outperforms baselines even though BigGAN was pretrained on ImageNet. 
Thanks to the rich representation power brought by CLIP,  
\our~can generate images 
with varying aspects,   
including artistic style, weather, background, texture, and etc., and is  capable of creating nonexistent,  counterfactual yet plausible objects.   
In addition, with the composed generation technique, we can 
generate better images with multiple objects. 
See Appendix for high resolution copies of the images shown in the paper.
%

\paragraph{Quantitative Evaluation on MS COCO Test Set}
To compare with other text-to-image generation methods, we evaluate our methods on a subset of 30,000 captions sampled from the COCO dataset. We follow the same standard evaluation protocol in~\cite{xu2018attngan, li2019object, tao2020df, zhu2019dm}, with the official code provided by~\cite{li2019object}\footnote{https://github.com/jamesli1618/Obj-GAN}.
We use Fréchet inception distance (FID),  Inception score (IS)
and R-precision to evaluate the performance.
For R-precision, 
following \cite{xu2018attngan, li2019object,zhu2019dm, tao2020df}, we compute the cosine similarity between a global image vector and 100 candidate sentence vectors, extracted by a pre-trained CNN-RNN retrieval model~\cite{xu2018attngan}. The candidate text descriptions include one ground-truth caption and 99 randomly selected unrelated sentences. R-precision is computed as the retrieval accuracy of all the 30,000 generated images.
We randomly repeat the process for 3 times and report the mean and the standard deviation of R-precision.
Note that baseline GANs are usually trained to maximize this score. For fair comparison, 
we replace the retrieval model used in~
\cite{xu2018attngan} with the CLIP text and image encoder, and report an additional CLIP R-precision score.

The results are shown in Table \ref{tab:coco}.
\our~achieves a comparable IS score to that of real images (34.26 versus 34.88).
Compared with DALL-E \cite{ramesh2021zero} and CogView \cite{ding2021cogview}, which are trained on billions of internet images with huge computation cost, 
we significantly improve the IS score from around 18 to 34, FID from 27 to 21 (e.g. FID 21.16 for {\our} with BigGAN-256, $k=5$).
Note that the BigGAN that we use was 
trained on ImageNet although the evaluation is on COCO images; we can expect to achieve better results by using a stronger generative model trained on COCO dataset.

\paragraph{Images Generated From COCO Captions}
We show a number of generated images given input captions from COCO dataset in Figure \ref{fig:coco}.
\our~ generates images with more details and objectives. For example, given \emph{`The traffic and people on a commercial street corner at night'}
, \our~can generates people, cars and a prosperous street with many lights. 

\paragraph{Varying Artistic Styles}
Although BigGAN is trained on the ImageNet  whose images are mostly realistic,  
with CLIP, {\our} is capable of producing meaningful images with different artistic styles, as displayed in the first row in Figure \ref{fig:background_texture}.  
The images are with six different styles, e.g. photo, monochrome, printmaking, painting, abstract painting and ink and wash painting. We can generate meaningful fake images with \emph{many granularities} even if the input sentence is complicated. 
Given the sentence (\emph{`old palaces and towers quivering within the wave's intenser day'}) from Percy Shelley's \textit{Ode to the West Wind}, \our~successfully generates palaces, towers, wave, and day light.

\paragraph{Varying Textures, Backgrounds and More}
As shown in \cite{harkonen2020ganspace,sauer2021counterfactual}, it is hard to control texture and background in standard GANs.
However, \our~can nicely control the texture and background of images through the input sentences. 
As shown in the second and third rows of Figure \ref{fig:background_texture}, 
{\our} easily put a car in different backgrounds (e.g. \emph{underwater}, \emph{night}, \emph{sky}) and with different textures (e.g. \emph{colorful glow}, \emph{starry}, \emph{ghost}).
Changing the object to robot, we can also generate meaningful but fake robots under different weather (e.g. \emph{rain}, \emph{sunshine}, \emph{snow}).
Moreover, \our~seems to show understanding about cultural difference by generating obviously different meals for the U.S., Russia and China: 
meal of \emph{the U.S.} contains corn, potato mesh and fried chicken; meal of \emph{Russia} contains black bread and Russian Borscht; meal of \emph{China} contains egg dumplings and spring rolls. 

\paragraph{Generating Counterfactual Contents}
In previous examples, we have shown some counterfactual examples, e.g. flaming dogs in Figure \ref{fig:intro_dogs}, car in the sky in Figure \ref{fig:background_texture}. 
Here, we use \our~to generate more 
high-quality counterfactual images with  different objects, background and style.

Figure \ref{fig:counterfactual} demonstrates that we can generate \emph{`glow and light dog'}, 
\emph{`castle in the sky'}, \emph{`cube butterfly'} and \emph{`underwater forest'}
. These images have different objects, backgrounds and styles, and do not appear in real world, nor in BigGAN's ImageNet training data.
It is surprising that {\our} successfully generates these out-of-domain images with high quality, especially given that we 
never change the parameters of the BigGAN.

\paragraph{Multiple Concepts with \ourcomp:}
We verify the performance of composed generation technique 
by generating images that contain two objects.
These two objects do not typically co-appear in normal images, e.g. cat and butterfly, dog and church, etc.
As shown in Figure \ref{fig:fusion}, {\our} (with single image generation) may entangle the two objects together or miss one of the objects. 
For example, \emph{`a dog near a boat'} gives a boat with a dog-like sail. \emph{`A butterfly near a boat'} generates only a butterfly while missing the boat.
However, by using composed generation, we can generate images with both objects. 
Even for more complicated sentences, we can generate meaningful and high-quality images (see the second row in Figure \ref{fig:fusion}).

To verify the robustness of our method to random seed, we  generate more images 
in Figure \ref{fig:multiple_images} for \emph{`A colorful robot walks on an iced lake under moonlight'}; we obtain a diverse set of images that are well related to the sentence.  


\section{Conclusions}

We propose {\our}, which enables high quality, state-of-the-art text-to-image generation with CLIP-guided GAN. 
Compared with traditional training-based approaches, 
our method is training-free, zero-shot, easily customizable, and is hence 
easily accessible to users with limited computational resource or special demands. 
Our novel techniques of {\ourloss} score, over-parameterized optimization and composed generation are of independent interest and useful in other latent space optimization problems.  

\newpage
{\small
\bibliographystyle{ieee_fullname}
\bibliography{egbib}
}

\clearpage
\appendix
\onecolumn
\section{Implementation Details}
We use the official pre-trained BigGAN model in PyTorch\footnote{https://github.com/ajbrock/BigGAN-PyTorch}. For initialization, we use $M=10,000$, where the batch size is $10$ and the initialization runs for $1,000$ steps. For optimization, we use Adam optimizer with a learning rate of $5\times10^{-3}$ with no weight decay, and optimize for $1,000$ iterations. On a GTX 3090 GPU, using BigGAN-256, \our~approximately requires 100 seconds for initialization and 80 seconds for optimization, resulting in 180 seconds ($\sim$3 minutes) in total. Changing the pre-trained model to BigGAN-512  increases the initialization time to 220 seconds and the optimization time to 120 seconds (yielding $\sim$7 minutes in total). 

\section{
More Studies on Composed Generation}
\begin{figure*}
    \centering
    \begin{subfigure}{1.\textwidth}
      \centering
      \includegraphics[width=.8\linewidth]{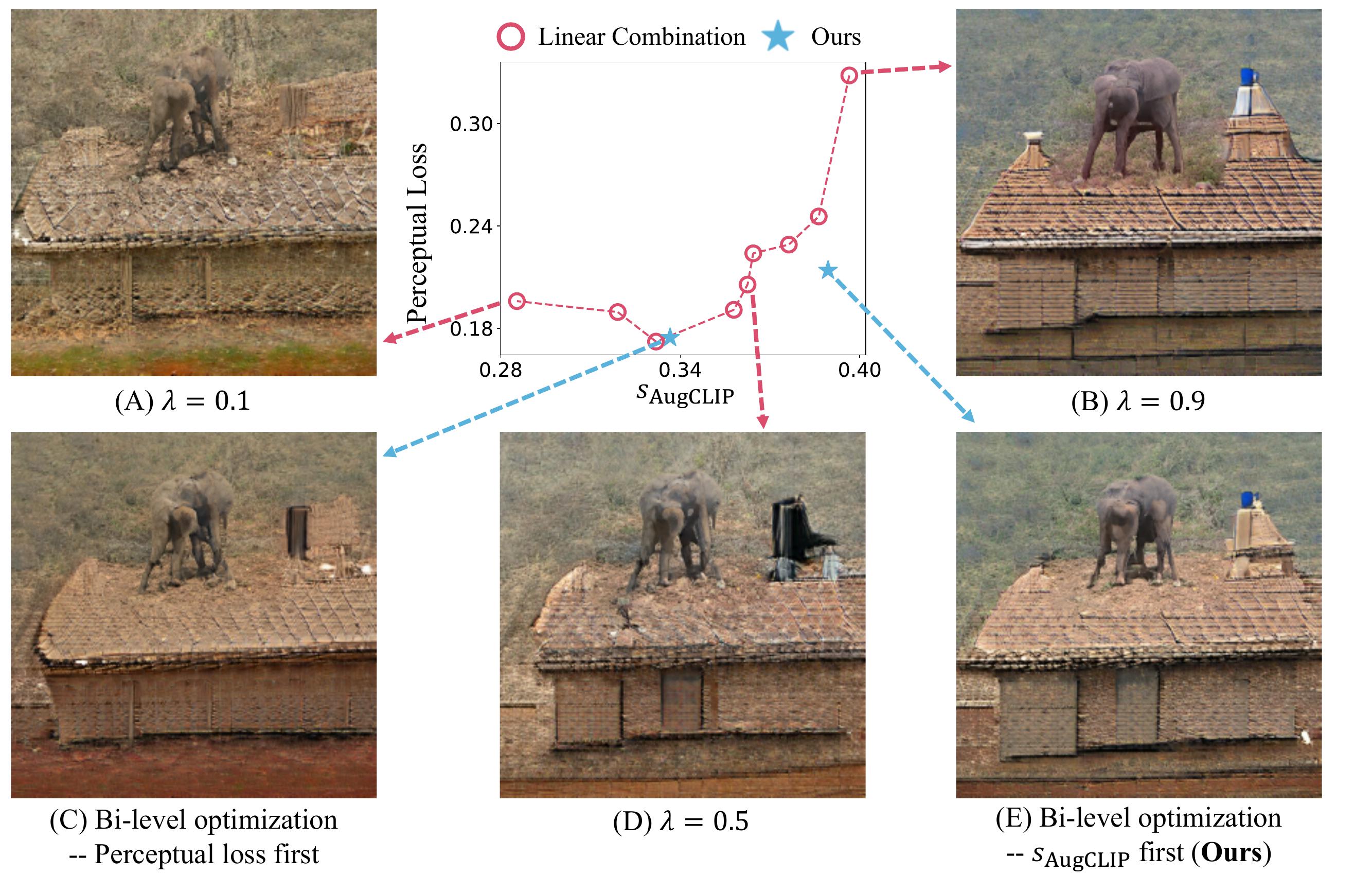}  
      \caption{Query Text: \emph{`An elephant on top of a roof.'}}
      \label{fig:lexico_elephant}
    \end{subfigure}
    
    \begin{subfigure}{1.\textwidth}
      \centering
      \includegraphics[width=.8\linewidth]{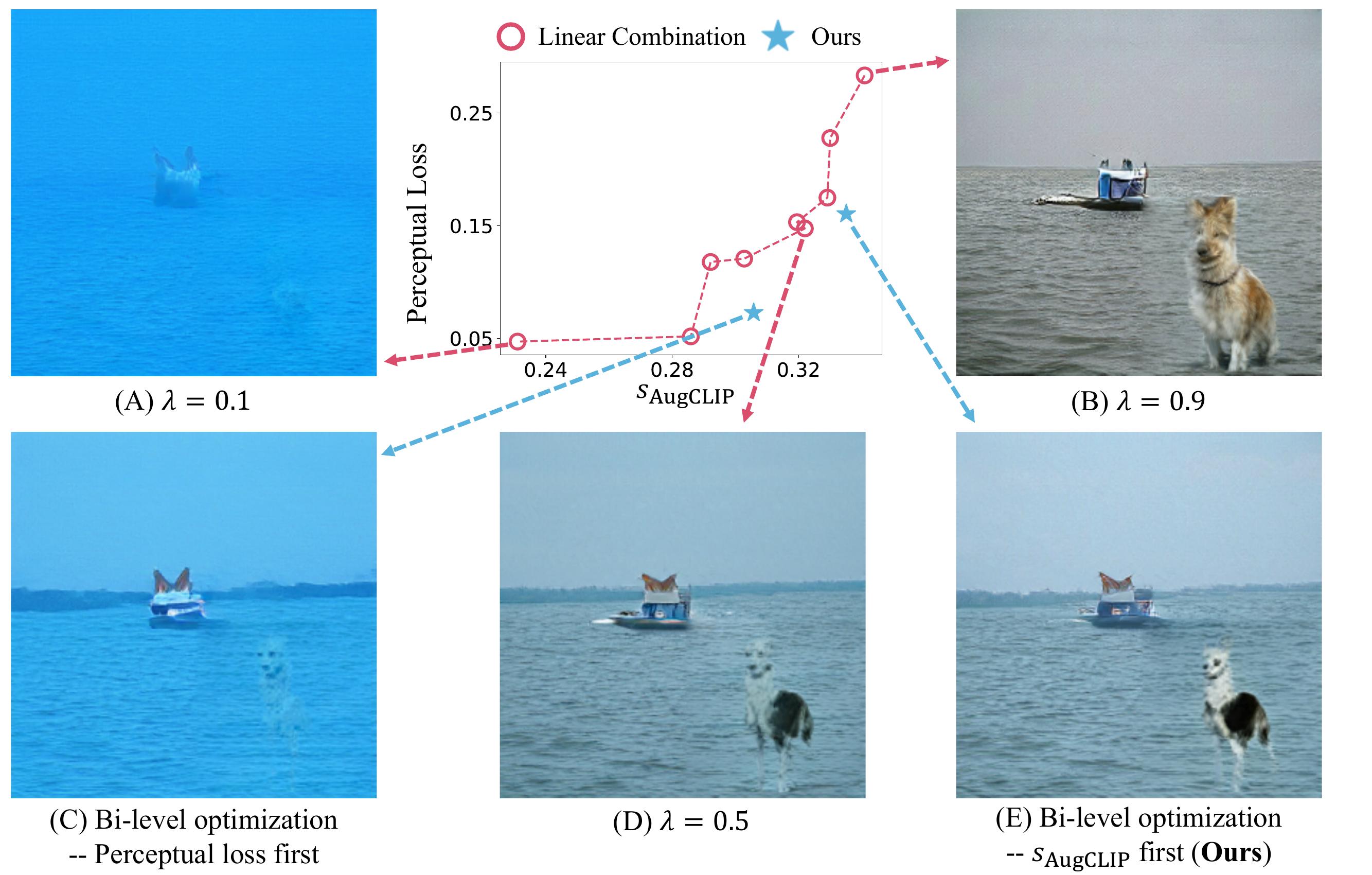}
      \caption{Query Text: \emph{`A dog near a boat'}}
      \label{fig:lexico_dog}
    \end{subfigure}
    \caption{We show two examples to demonstrate the benefit of using our bi-level optimization formulation in composed generation. We compare our formulation~\eqref{eq:bilevel} with the inverse bi-level optimization~\eqref{eq:r-bilevel} and linear combination method~\eqref{eq:linear}. We search the linear combination coefficient $\lambda$ from $0.1$ to $0.9$ uniformly. Observations: 
    (1) For the linear combination method, 
    For two text queries (a) and (b), the effect of similar $\lambda$ is different. For instance, when $\lambda = 0.5$, the generated image for (a) has acceptable visual quality, but the generated image for (b) failed to generate recognizable \emph{`dog'}; (2) Instead, our bi-level optimization formulation~\eqref{eq:bilevel} with dynamic-barrier gradient descent relieves the user from tuning $\lambda$ for each image. It yields low perceptual loss without sacrificing too much of the \ourloss~score, and finally generates natural fused images; (3) The inverse bi-level optimization~\eqref{eq:r-bilevel} formulation cannot get good results.}
    \label{fig:appendix_lexico}
\end{figure*}

We discuss the design choices for the composed generation in Sec.~\ref{subsec:fusion}. 
To demonstrate the effectiveness of the bi-level optimization,  formulation~\eqref{eq:bilevel}, 
we consider 
two alternative formulations
for trading off the \ourloss~score $s_{\texttt{Fuse}}$ and the perceptual loss $\ell_{\texttt{Fuse}}$  of the fused image:  \vspace{.5\baselineskip}

\paragraph{Linear Combination} 
A standard way to trade-off two loss functions is to optimize their linear combination: 
\begin{equation}
\label{eq:linear}
\begin{aligned}
    & \min_{\bar{\vv \xi}, \bar{\vv\alpha}}~~~~(1-\lambda)\ell_{\fuse}(\bar{\vv \xi}, \bar {\vv \alpha}) - \lambda s_{\fuse}(\bar{\vv \xi}, \bar {\vv \alpha}),
\end{aligned}
\end{equation}
where $\lambda \in(0,1)$ is a linear combination coefficient used to balance the two objectives. 
However, as shown in Figure~\ref{fig:appendix_lexico}, 
the key disadvantage of this approach is that  
the optimal choice of $\lambda$ depends on 
the query text, and hence needs to be tuned by the user 
case by case. This makes the overall procedure  computationally expensive and difficult to automatize.  
In comparison, the bi-level optimization approach does not require to tune $\lambda$ case by case, and provides 
high quality fused image with only a single run. 

\paragraph{Inverse Bi-level Optimization}
In the bi-level optimization in \eqref{eq:bilevel}, 
we prioritize the optimization of the 
 \ourloss~score $s_{\texttt{Fuse}}$ while adding the perceptual loss  $\ell_{\texttt{Fuse}}$  as the secondary loss. 
 An alternative approach is to switch the roles of the two loss functions, 
prioritizing $\ell_{\texttt{Fuse}}$  and treating $s_{\texttt{Fuse}}$ as the seondary loss: 
\begin{equation}
\label{eq:r-bilevel}
\begin{aligned}
    & \max_{\bar{\vv \xi}, \bar{\vv\alpha}}  s_{\fuse}(\bar{\vv \xi}, \bar {\vv \alpha})
    & s.t. ~~~ (\bar{\vv \xi}, \bar{\vv \alpha}) \in \argmin \ell_{\fuse}(\bar{\vv \xi}, \bar {\vv \alpha}).
\end{aligned}
\end{equation}
As shown in Figure.~\ref{fig:appendix_lexico},
this approach does not work as well as  \eqref{eq:bilevel}, 
because it tends to generate images with poor \ourloss~score. 
Intuitively, \eqref{eq:bilevel} is better because 
$s_{\texttt{Fuse}}$ is difficult to optimize and  
$\ell_{\texttt{Fuse}}$ is much easier to optimize, and hence it makes more sense to prioritize the optimization of $s_{\texttt{Fuse}}$. 
\vspace{.5\baselineskip}

\section{Additional Experimental Results}

\begin{figure*}
    \centering
    \includegraphics[width=1.\textwidth]{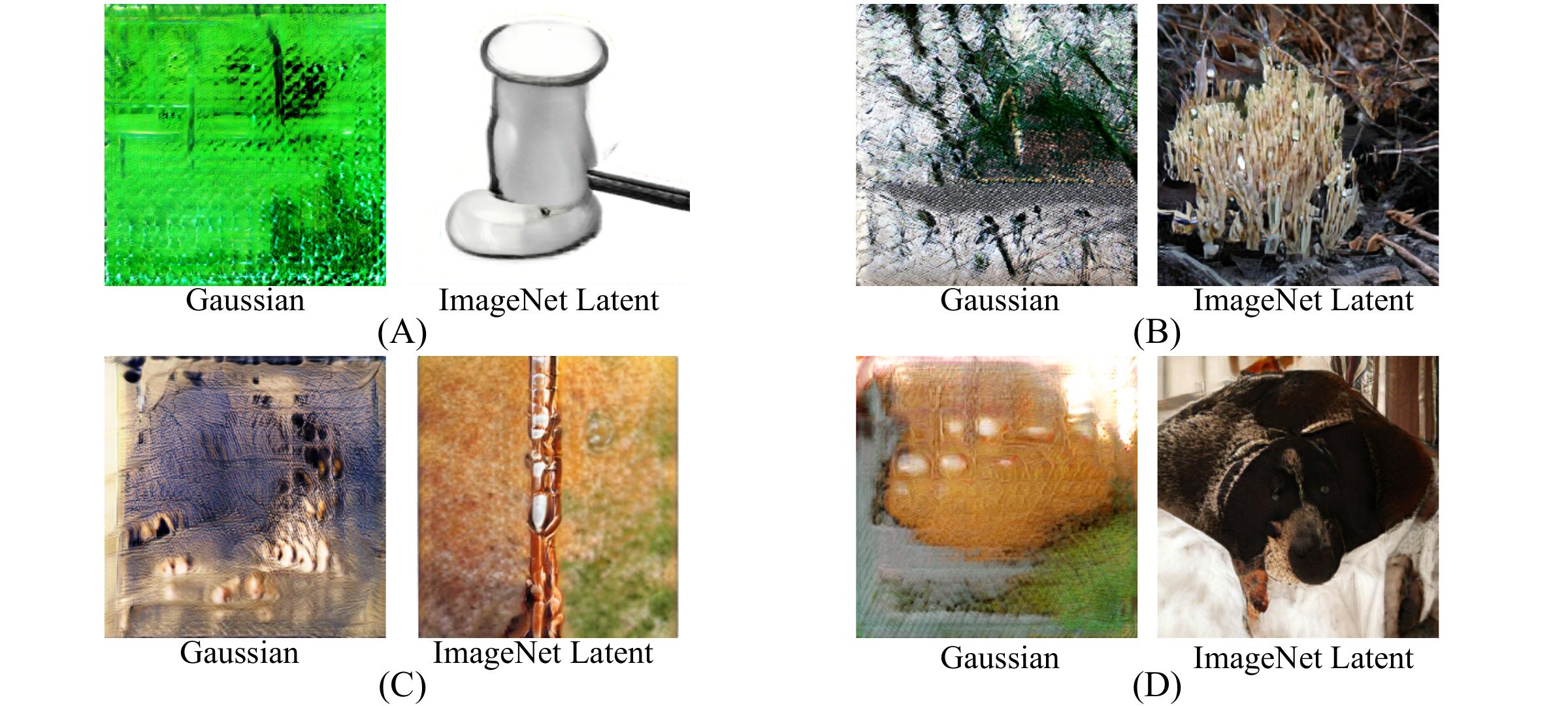}
    \caption{This figures show the other initialization choice (initialize $\vec z$ and $\vec y$ from the standard Gaussian distribution $\mathcal{N}(0, I)$) for generating the images in Figure.~\ref{fig:opt}, as mentioned in the main text. After optimization, the generated images are still noises. In comparison, initializing from the latent codes of ImageNet classes gives more natural images. The corresponding query texts are:~(A) This small bird has a pink breast and crown, and black primaries and secondaries. (B) A church near forest under moonlight. (C) A photo of an ice cube melting under the sun. (D) An armchair in the shape of an avocado. 
    }
    \label{fig:noise_init_opt}
\end{figure*}

\paragraph{Choice of Initialization} As we discussed in  Section~\ref{subsec:strategy}, 
it is recommended to initialize the class label $\vv y$ by randomly selecting from the latent representations for the 1,000 classes of ImageNet 
. 
An alternative approach is to initialize $\vec y$ from the standard Gaussian distribution, 
which, however, leads to highly noisy images as we show in Figure.~\ref{fig:noise_init_opt}. 
\vspace{.5\baselineskip}

\paragraph{Linear Interpolation} We linearly interpolate between two generated images to examine the intermediate images in the latent space.
Specifically, 
given two query text $\mathcal T_1$ and $\mathcal T_2$, 
and let $\vv \xi_1$ and $\vv \xi_2$ be the latent code provided by FuseDream (without composed generation), 
 we generate a sequence of images via 
 $$\mathcal I_\alpha = 
 g(\alpha\vv \xi_1 + (1-\alpha)\vv \xi_2 ), 
 $$
 where $\alpha \in[0,1]$. 
As shown in Figure.~\ref{fig:linear_interp}, 
the intermediate images provides a smooth 
interpolation between the images of the two queries. 
\vspace{.5\baselineskip}

\begin{figure*}
\centering
\includegraphics[width=1.\textwidth]{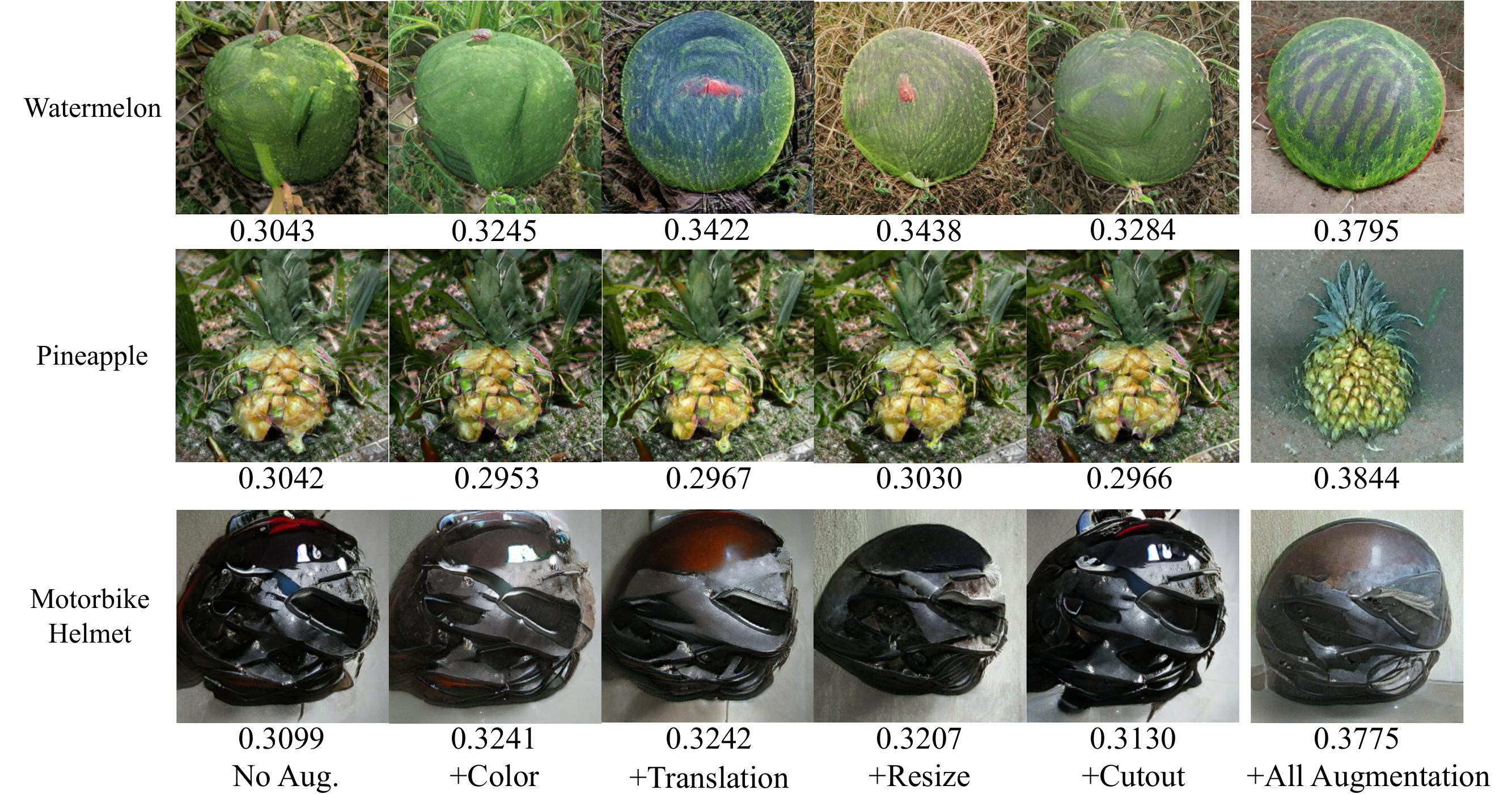}
\caption{
Images generated by {\our} when we incorporate different data augmentation techniques in {\ourloss}. 
The numbers indicate their corresponding \ourloss~score computed with all data augmentation. We observe that each data augmentation technique has their unique influence on the generated image, but the best visual result is obtained by applying all the augmentation techniques.}
\label{fig:aug}
\end{figure*}

\paragraph{Influence of the Data Augmentation Techniques} We perform an ablation study on choice of  data augmentation techniques (random colorization, random translation, random resize, and random cutout) in the {\ourloss} score.  Results are shown in Figure.~\ref{fig:aug}. 

\vspace{.5\baselineskip}

\paragraph{Out-of-domain Generation} We use query texts including famous landmarks, arts, animation figures, etc., to examine the ability of \our~to generate images outside the training domain of ImageNet. 
As shown in Figure.~\ref{fig:aug}, 
{\our} can generate famous landmarks, masterpieces, emojis, cartoon characters that 
are not included in the ImageNet training data. 
\vspace{.5\baselineskip}

\begin{figure*}
\centering
\includegraphics[width=1.\textwidth]{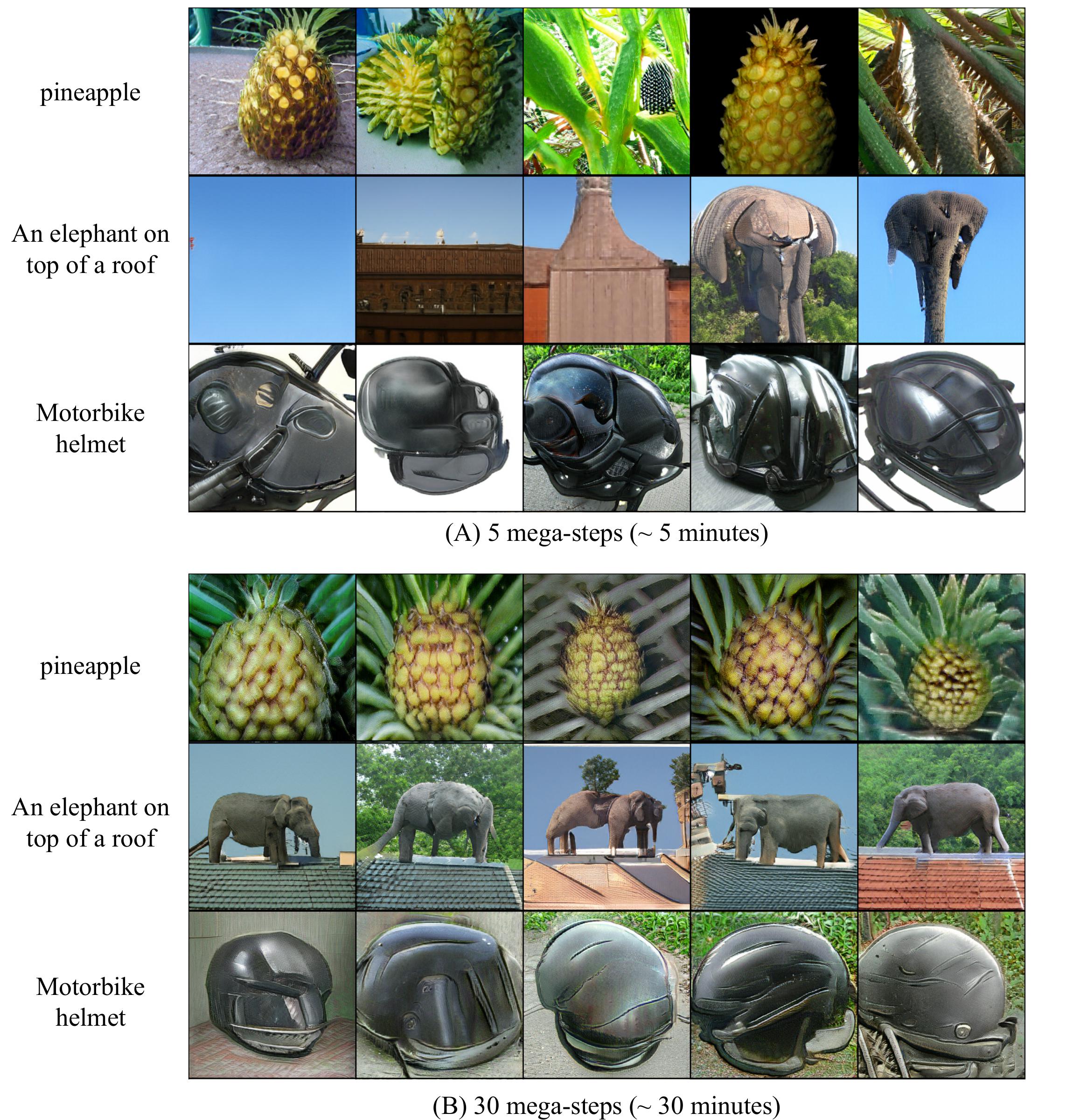}
\caption{Direct optimization of the  \ourloss~score with BasinCMA~\cite{wampler2009optimal} without our proposed initialization and over-parameterization strategy. `Mega-step' refers to the number of iterations in the outer loop when using BasinCMA. We use the same computer with GTX 3090 to test the computational time. The base GAN is BigGAN-256. The objective is $s_{\ourloss}$. In Figure.(A), with similar running time ($\sim$5 minutes), BasinCMA failed to generate realistic images for `pineapple' and `motorbike helmet', while \our~successes (See Figure.~\ref{fig:aug}). In Figure.(B), with longer optimization time ($\sim$30 minutes), BasinCMA can generate semantically related images.}
\label{fig:basincma}
\end{figure*}

\paragraph{Optimization with Gradient-free Optimizer}
We replace Adam and optimize \ourloss~score with BasinCMA optimizer~\cite{wampler2009optimal}, which is a gradient-free optimizer used in previous works~\cite{huh2020transforming, bau2019seeing} for optimizing in the GAN latent space. Results are shown in Figure.~\ref{fig:basincma}.
\vspace{.5\baselineskip}

\begin{figure*}[t]
\centering
\scalebox{0.96}{
\begin{tabular}{c}
\includegraphics[width=1.\textwidth]{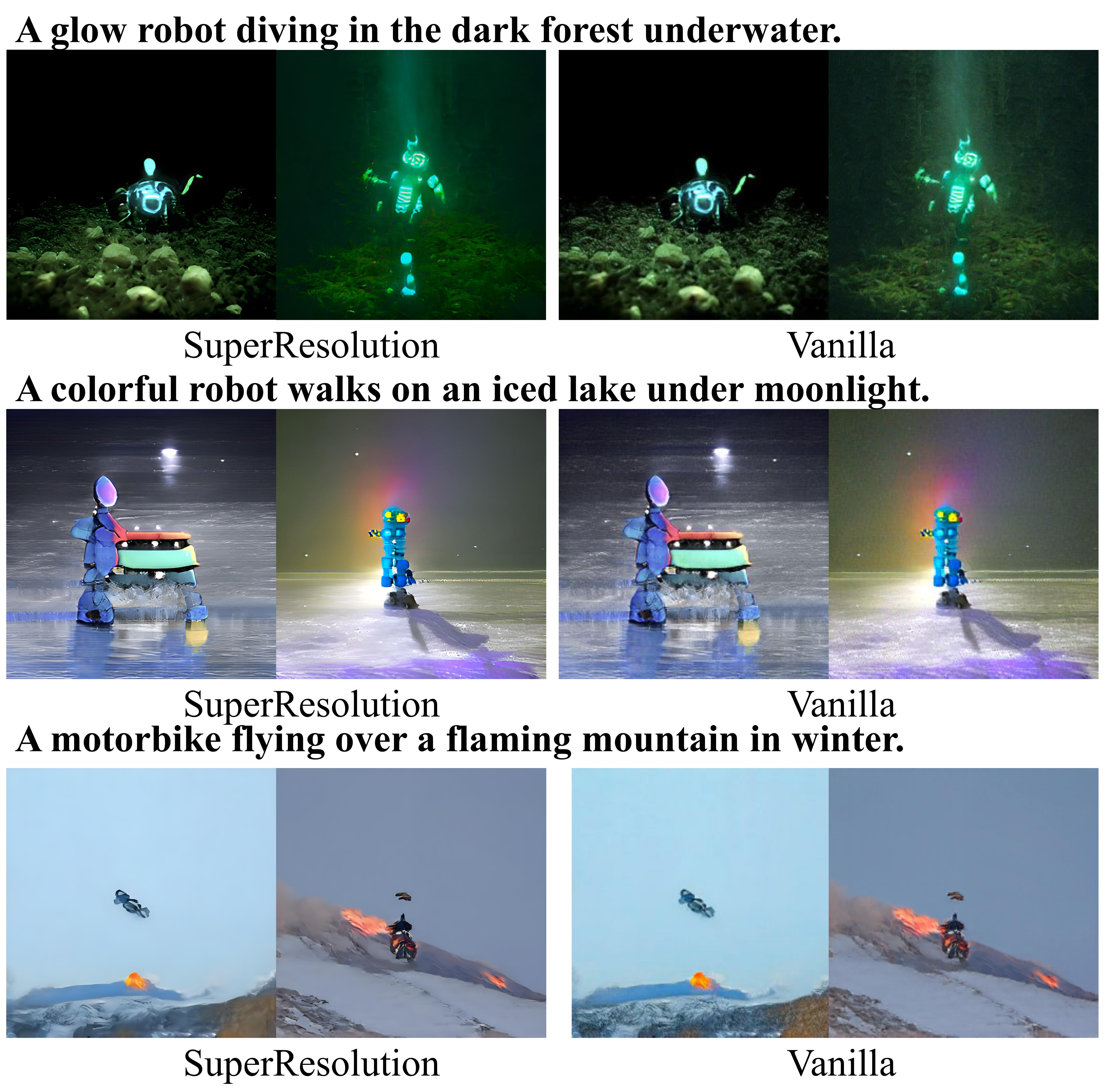}
\\
\end{tabular}}
\caption{
Original images and their super-resolution version ($\times~4$ ) generated by \cite{wang2021realesrgan}.
}
\label{fig:fusionHR}
\end{figure*}

\paragraph{High-resolution images} We use~\cite{wang2021realesrgan} to get high-resolution version of sampled generated images in Figure \ref{fig:fusionHR}.

\begin{figure*}
    \centering
    \includegraphics[width=1.\textwidth]{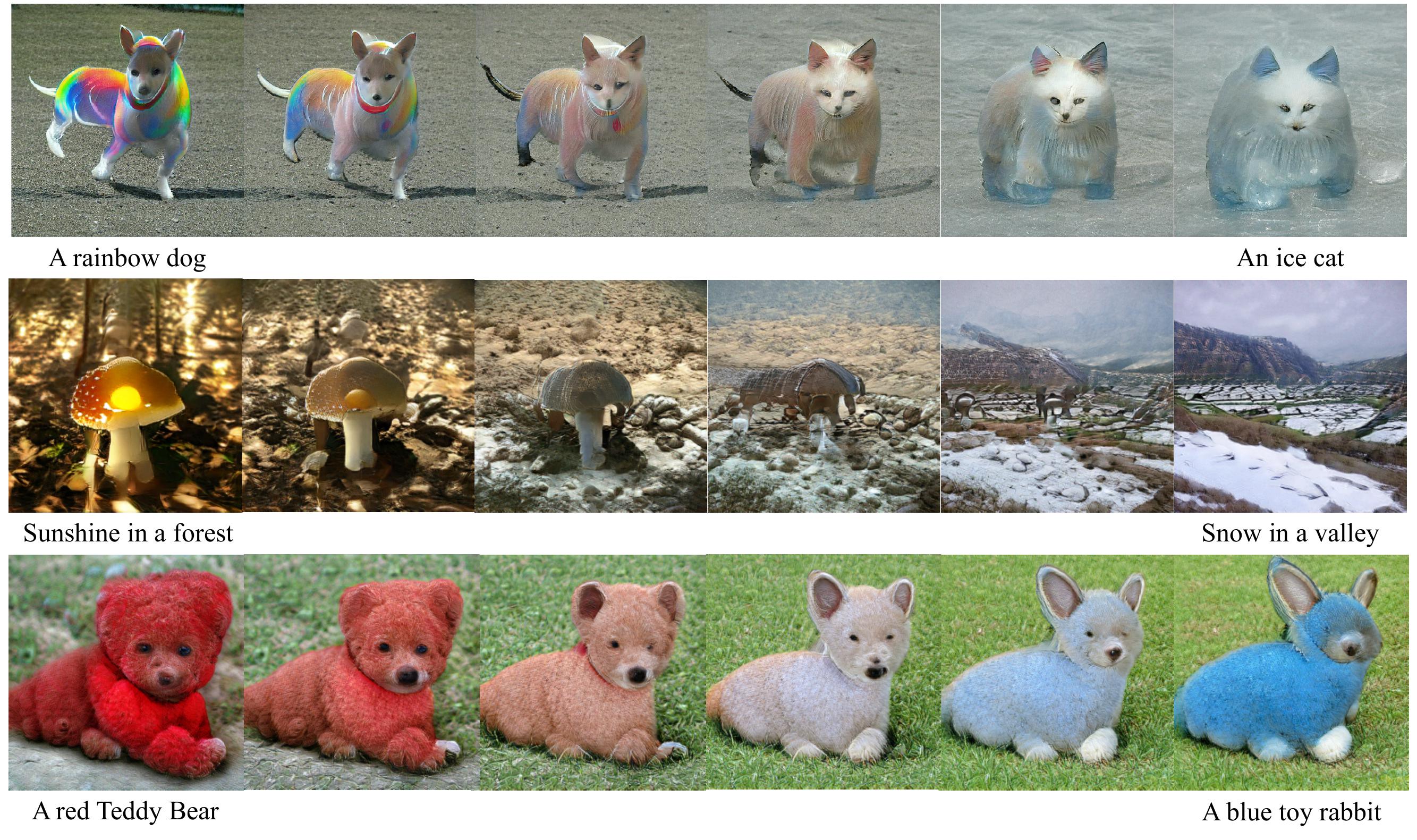}
    \caption{Three examples of linearly interpolating between the latent codes of two generated images. We observe a smooth transition and the intermediate results are still natural and realistic. 
    }
    \label{fig:linear_interp}
\end{figure*}

\begin{figure*}
    \centering
    \includegraphics[width=1.\textwidth]{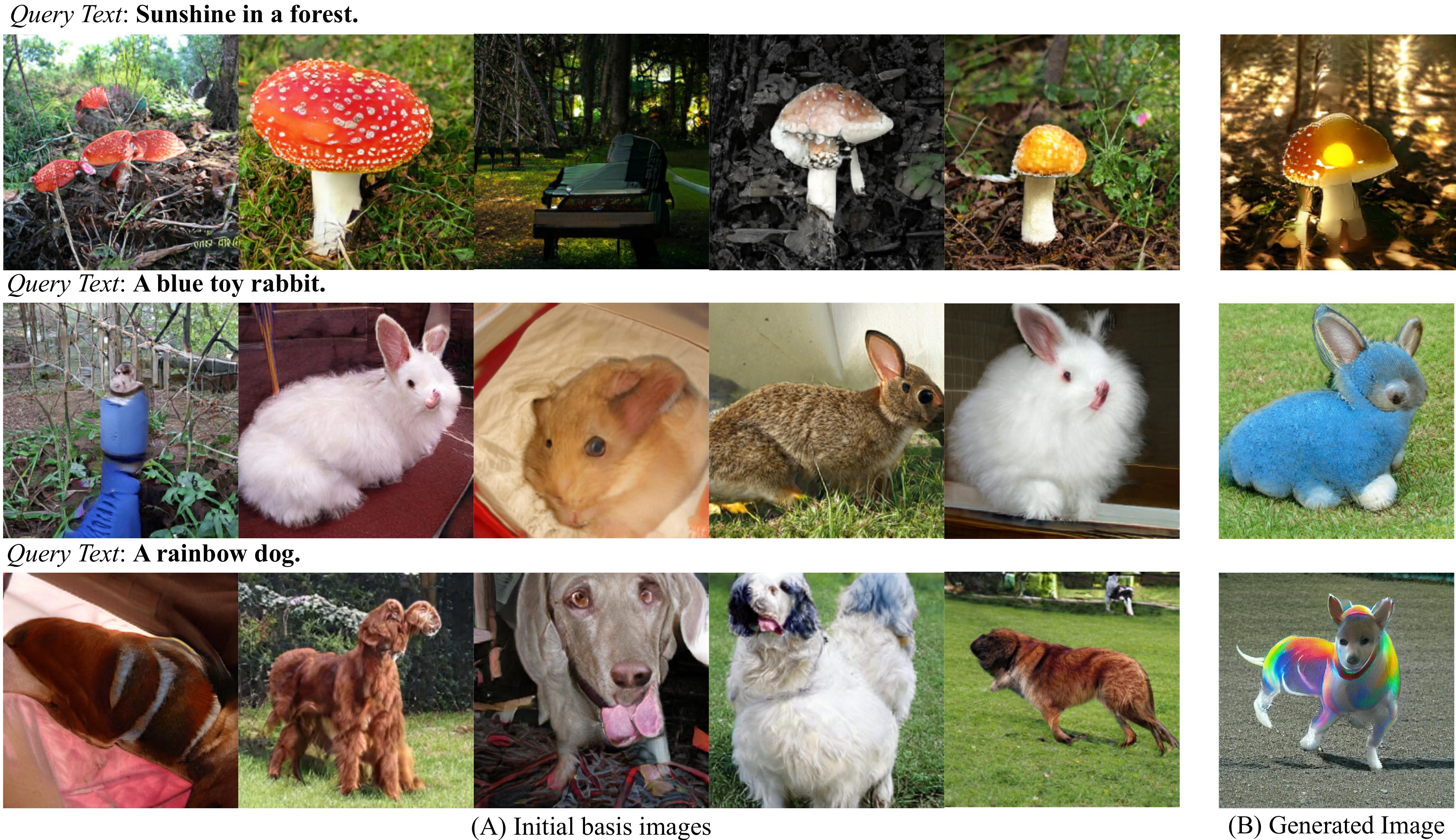}
    \caption{The initial basis images when $k=5$ for the query text \emph{`Sunshine in a forest'} (top row), \emph{`A blue toy rabbit'} (middle row) and \emph{`A rainbow dog'} (bottom row).} 
    \label{fig:linear_interp}
\end{figure*}

\begin{figure*}
\centering
\includegraphics[width=1.\textwidth]{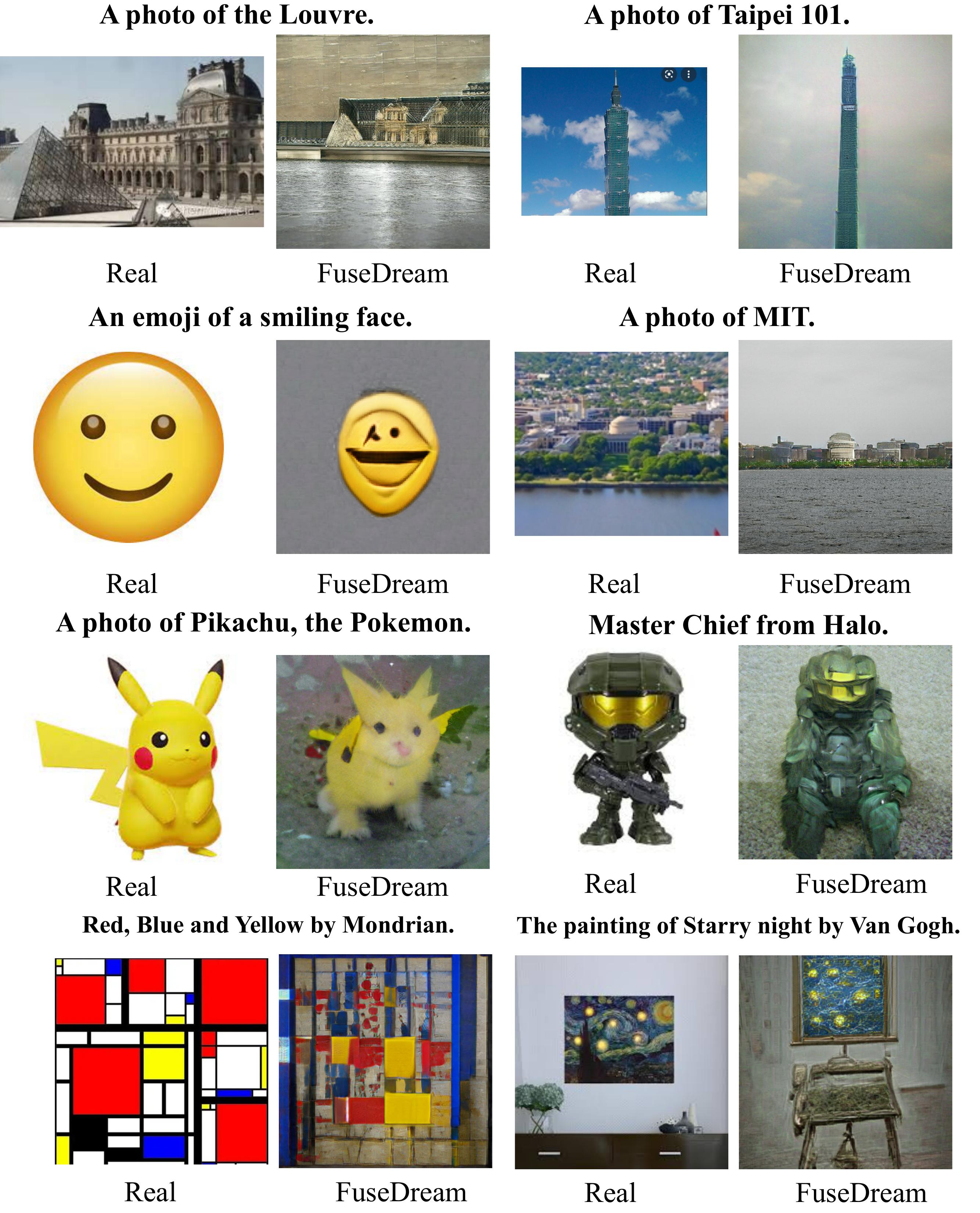}
\caption{More examples for demonstrating the ability of $\our$ to generate images that are outside the training domain (ImageNet) of BigGAN. The `real' images are adopted from the Internet for comparison.
}
\label{fig:famous}
\end{figure*}

\end{document}